%% file: main.tex
\documentclass{article}

\usepackage[utf8]{inputenc} %
\usepackage[T1]{fontenc}    %
\usepackage{hyperref}
\usepackage{url}
\usepackage{amsfonts}       %
\usepackage{nicefrac}       %
\usepackage{microtype}      %

\usepackage{amsmath}
\usepackage{amssymb}
\usepackage{mathtools}
\usepackage{amsthm}
\usepackage{algorithm}
\usepackage{algpseudocode}

\usepackage{multirow}
\usepackage{booktabs}
\usepackage{subcaption}

\usepackage{graphicx}
\graphicspath{{images/}}

\usepackage{enumitem}
\usepackage{xcolor}
\usepackage{colortbl}
\usepackage{pifont}
\usepackage{wrapfig}

\input{math_commands.tex}

\usepackage[capitalize,noabbrev]{cleveref}

\newcommand{\ADM}{\texttt{ADM}}

\newcommand{\Cor}{\mathrm{Cor}}
\newcommand{\dCor}{\mathrm{dCor}}
\newcommand{\mCor}{\mathrm{mCor}}

\makeatletter
\renewcommand{\paragraph}{%
  \@startsection{paragraph}{4}%
  {\z@}{0em}{-0.5em}%
  {\normalfont\normalsize\bfseries}%
}
\makeatother

\newcommand{\indep}{\perp \!\!\! \perp}

\definecolor{lightgray}{HTML}{EFEFEF}

\usepackage[preprint]{icml2026}

\usepackage[capitalize,noabbrev]{cleveref}

\theoremstyle{plain}
\newtheorem{theorem}{Theorem}[section]

\newtheorem{corollary}[theorem]{Corollary}
\theoremstyle{definition}

\newtheorem{example}{Example}
\theoremstyle{remark}

\icmltitlerunning{Adversarial Dependence Minimization}

\begin{document}

\twocolumn[
  \icmltitle{Adversarial Dependence Minimization}

  \icmlsetsymbol{equal}{*}

  \begin{icmlauthorlist}
    \icmlauthor{Pierre-François De~Plaen}{esat}
    \icmlauthor{Tinne Tuytelaars}{esat}
    \icmlauthor{Marc Proesmans}{esat}
    \icmlauthor{Luc Van~Gool}{esat,cvl,insait}
  \end{icmlauthorlist}

  \icmlaffiliation{esat}{ESAT-PSI, KU Leuven, Belgium}
  \icmlaffiliation{cvl}{CVL, ETH Zürich, Switzerland}
  \icmlaffiliation{insait}{INSAIT, Sofia University, Bulgaria
    }

  \icmlcorrespondingauthor{Pierre-François De~Plaen}{pdeplaen@esat.kuleuven.be}

  \icmlkeywords{representation learning, decorrelation, independent, adversarial, PCA, ICA, generalization, SSL}

  \vskip 0.3in
]

\printAffiliationsAndNotice{}  %

\begin{abstract}
    Minimally redundant representations are typically learned by minimizing feature covariance.
    However, covariance-based methods fail to eliminate all dependencies/redundancies, as linearly uncorrelated variables can still exhibit nonlinear relationships. 
    To address this, we introduce \ADM, a differentiable algorithm that minimizes statistical dependence between feature dimensions through an adversarial game: auxiliary networks identify dependencies, while the encoder removes them. 
    We prove that mutual independence is achieved at the global optimum, empirically verify convergence, and study three potential applications: 
    extending PCA to nonlinear decorrelation, improving generalization in image classification, and preventing dimensional collapse in self-supervised learning. 
    By promoting statistically independent representations, \ADM{} paves the way for learning more robust, compressed, and generalizable representations across diverse applications.
\end{abstract}

\section{Introduction}
\input{sections/introduction}

\section{Background}
\input{sections/background}

\section{Related Work}
\input{sections/related_work}

\section{Adversarial Dependence Minimization}
\input{sections/method}

\section{Applications}
\input{sections/applications}

\section{Experiments}
\input{sections/experiments}

\section{Open Questions and Future Directions}
\input{sections/forward}

\section{Conclusion}
\input{sections/conclusion}

\bibliography{bib}
\bibliographystyle{icml2026}

\newpage
\appendix
\onecolumn

\input{appendix/proof_theroem1}
\input{appendix/pca_extension}
\input{appendix/extended_related}

\input{appendix/independence}

\input{appendix/algo}
\input{appendix/ablation_studies}

\input{appendix/additional_results}
\input{appendix/training_details}

\end{document}

%% file: math_commands.tex
\usepackage{amsfonts,bm}  %

\def\eqref#1{equation~\ref{#1}}

\def\floor#1{\lfloor #1 \rfloor}
\def\1{\bm{1}}

\def\ru{{\textnormal{u}}}
\def\rv{{\textnormal{v}}}

\def\rx{{\textnormal{x}}}

\def\rz{{\textnormal{z}}}

\def\rvv{{\mathbf{v}}}

\def\rvx{{\mathbf{x}}}

\def\rvz{{\mathbf{z}}}

\def\rmX{{\mathbf{X}}}

\def\vtheta{{\bm{\theta}}}
\def\vphi{{\bm{\phi}}}
\def\vpsi{{\bm{\psi}}}

\def\vb{{\bm{b}}}

\def\vx{{\bm{x}}}

\def\vz{{\bm{z}}}

\def\evx{{x}}

\def\mI{{\bm{I}}}

\def\mW{{\bm{W}}}
\def\mX{{\bm{X}}}

\DeclareMathAlphabet{\mathsfit}{\encodingdefault}{\sfdefault}{m}{sl}
\SetMathAlphabet{\mathsfit}{bold}{\encodingdefault}{\sfdefault}{bx}{n}

\newcommand{\R}{\mathbb{R}}

\newcommand{\softmax}{\mathrm{softmax}}

\newcommand{\Cov}{\mathrm{Cov}}

%% file: sections/introduction.tex
\begin{figure}[t!]
    \centering
    \def\svgwidth{0.98\columnwidth}
    {\fontsize{8.5pt}{9.5pt}\selectfont 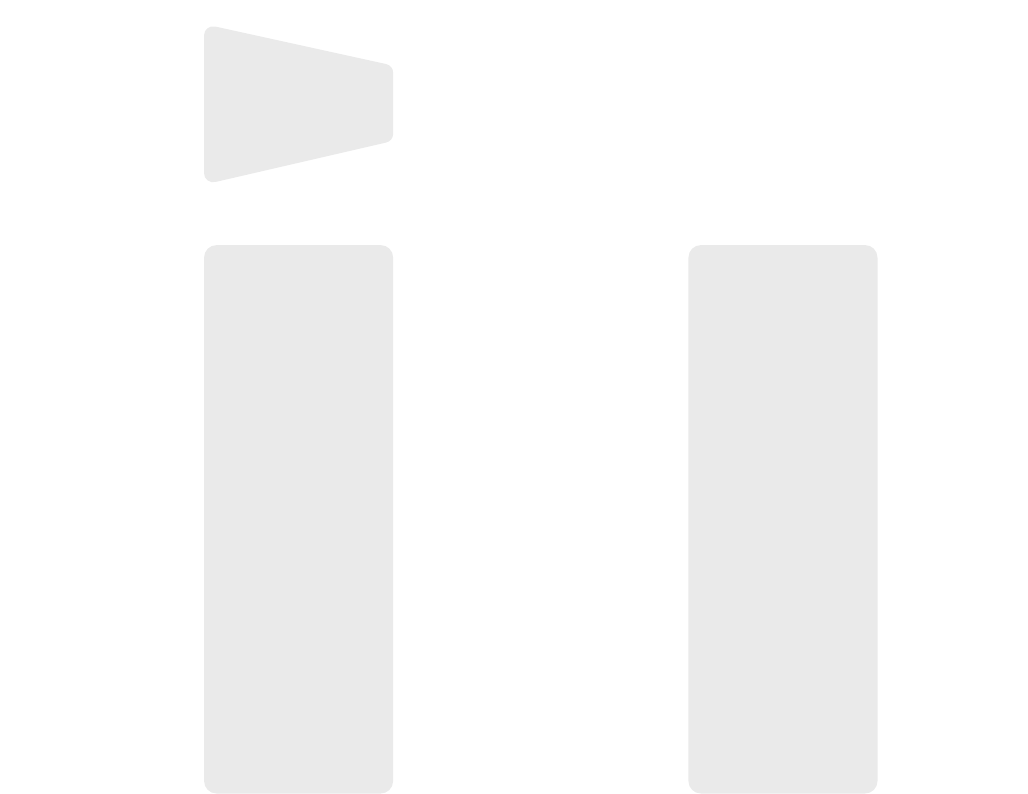}
    \caption{Overview of Adversarial Dependence Minimization. Probe networks expose dependencies among latent dimensions, while predictors reconstruct each transformed dimension from the others. The encoder adversarially maximizes the prediction error, driving the representation toward statistical independence.}
    \label{fig:adversarial_game_illustr}
\end{figure} 

In representation learning~\citep{rumelhart1986learning_repres,hinton2006deep_belief,bengio2013representation_learn}, algorithms learn to extract lower-dimensional encodings from input data, aiming for compact yet informative representations. 
A key strategy is to learn representations with minimally redundant dimensions, where each feature encodes a distinct concept. 
Such minimally redundant dimensions may offer several advantages, including efficient data compression, compositional generalization, and enhanced interpretability of the learned representations.

\begin{figure*}
    \vspace{-8px}
    \centering
    \begin{subfigure}[t]{0.31\textwidth}  
        \centering
        \def\svgwidth{120px}
        {\scriptsize 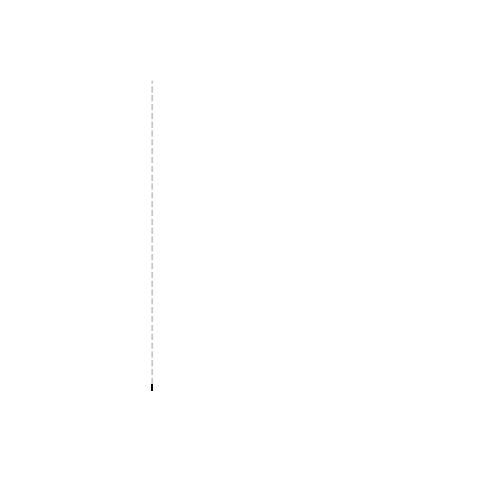}
        \vspace{-4px}
        \caption{Lin. $\rx_2=-\rx_1$: $\Cor = -1$ and $\dCor = 1$.}
    \end{subfigure}
    \hfill
    \begin{subfigure}[t]{0.35\textwidth}
        \centering
        \def\svgwidth{120px}
        {\scriptsize 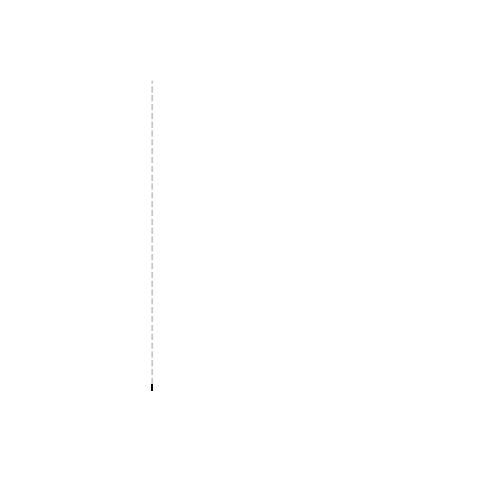}
        \vspace{-4.5px}
        \caption{Nonlin. $\rx_2=\rx_1^2$: $\Cor = 0$ and \smash{$\dCor = \sqrt{1/2}$}.}
        \label{subfig:limitation_of_linear}
    \end{subfigure}
    \hfill
    \begin{subfigure}[t]{0.31\textwidth}
        \centering
        \def\svgwidth{120px}
        {\scriptsize 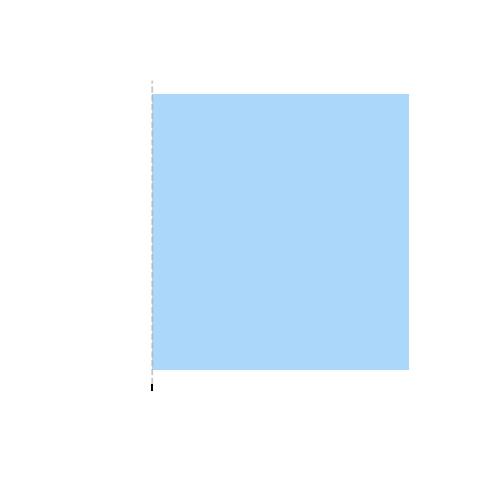}
        \vspace{-4px}
        \caption{Indep. $\rx_2 \in \mathcal{U}(-1,1)$: $\Cor = \dCor = 0$.}
    \end{subfigure}
    \caption{Illustration of the joint and marginal distributions for different types of dependencies between random variables $\rx_1 \in \mathcal{U}(-1,1)$ and $\rx_2$. The variables are linearly uncorrelated ($\Cor = 0$) in both (b) and (c) but are independent only in (c).} 
    \label{fig:correlation_and_dependence_ex}
\end{figure*}

The strategy of minimizing feature redundancy has been applied in various methods, from classical techniques like the Principal Component Analysis~\citep{pearson1901pca_lines_and_planes, hotelling1933pca_2} to more recent self-supervised learning (SSL) approaches~\citep{huang2018decorrelated_bn, zbontar2021barlow_ssl, bardes2021vicreg_ssl}. Both PCA and these SSL methods aim to extract representations with uncorrelated dimensions. 
However, it is important to note that while these methods minimize pairwise linear correlations ($\Cor$) between embedding dimensions, non-linear dependencies may still be present. 
We illustrate this with Example~\ref{example:uncorrelated_but_dependent} (Figure~\ref{subfig:limitation_of_linear}). 
\begin{example} \label{example:uncorrelated_but_dependent}
    Let a random variable $\rx_1$ be drawn from a uniform distribution over the interval $\left[-a,a\right]$ and $\rx_2 = \rx_1^2$. Since $\rx_2$ is a deterministic function of $\rx_1$, it is clear that the variables co-vary. 
    However, we find that: $\Cor(\rx_1, \rx_2) = 0$. 
    The random variables are thus dependent despite a zero Pearson correlation. 
\end{example} 
This example highlights a limitation of algorithms that rely on Pearson correlation: they may not eliminate all forms of dependencies/redundancies, as linearly uncorrelated variables can still exhibit nonlinear relationships~\citep{hyvarinen2000ICA_book}. Those methods may thus learn redundant embeddings, in which multiple dimensions encode the same concept. 
Still, developing a stable method for mutual and nonlinear decorrelation remains an unresolved problem. The challenge lies in designing a training objective that is simultaneously differentiable, scalable, and distribution-agnostic.

This paper presents a training algorithm for learning embeddings with minimal dependence among dimensions, achieved by removing linear, nonlinear, and higher-order relationships.  
The method is formulated as a minimax adversarial game between two types of players: 
(1) a series of small neural networks is trained to learn how embedding dimensions relate, and (2) an encoder is trained to counter them by updating the representations. 

Our adversarial objective can be interpreted as a soft independence constraint in a task-specific optimization problem, enabling its incorporation into various learning algorithms. In this work, we explore three potential applications:
\begin{enumerate}[itemsep=2pt,topsep=0pt,parsep=0pt,partopsep=0pt]
    \item extending PCA to non-linear decorrelation
    \item learning features that generalize beyond label supervision in supervised learning
    \item preventing dimensional collapse in SSL by learning minimally redundant representations
\end{enumerate}

This paper's contributions are: (1) the introduction of an algorithm that minimizes all forms of dependencies between learned features, (2) proving that dimensions are statistically independent at the optimal solution, (3) showing empirically that the algorithm converges to this equilibrium, and (4) studying three potential applications.

%% file: images/ADM_illustr_v2b.pdf_tex
\begingroup%
  \makeatletter%
  \providecommand\color[2][]{%
    \errmessage{(Inkscape) Color is used for the text in Inkscape, but the package 'color.sty' is not loaded}%
    \renewcommand\color[2][]{}%
  }%
  \providecommand\transparent[1]{%
    \errmessage{(Inkscape) Transparency is used (non-zero) for the text in Inkscape, but the package 'transparent.sty' is not loaded}%
    \renewcommand\transparent[1]{}%
  }%
  \providecommand\rotatebox[2]{#2}%
  \newcommand*\fsize{\dimexpr\f@size pt\relax}%
  \newcommand*\lineheight[1]{\fontsize{\fsize}{#1\fsize}\selectfont}%
  \ifx\svgwidth\undefined%
    \setlength{\unitlength}{493.01449585bp}%
    \ifx\svgscale\undefined%
      \relax%
    \else%
      \setlength{\unitlength}{\unitlength * \real{\svgscale}}%
    \fi%
  \else%
    \setlength{\unitlength}{\svgwidth}%
  \fi%
  \global\let\svgwidth\undefined%
  \global\let\svgscale\undefined%
  \makeatother%
  \begin{picture}(1,0.77257935)%
    \lineheight{1}%
    \setlength\tabcolsep{0pt}%
    \put(0,0){\includegraphics[width=\unitlength,page=1]{ADM_illustr_v2b.pdf}}%
    \put(0.45808391,0.74707236){\makebox(0,0)[lt]{\lineheight{1.25}\smash{\begin{tabular}[t]{l}Repres. z\end{tabular}}}}%
    \put(0.6590567,0.66188217){\makebox(0,0)[lt]{\lineheight{1.25}\smash{\begin{tabular}[t]{l}Task-specific loss\end{tabular}}}}%
    \put(0.48933045,0.00470068){\makebox(0,0)[lt]{\lineheight{1.25}\smash{\begin{tabular}[t]{l}MSE\end{tabular}}}}%
    \put(0.03052041,0.6467457){\makebox(0,0)[lt]{\lineheight{1.25}\smash{\begin{tabular}[t]{l}x\end{tabular}}}}%
    \put(-0.00343396,0.6852182){\makebox(0,0)[lt]{\lineheight{1.25}\smash{\begin{tabular}[t]{l}Input\end{tabular}}}}%
    \put(0,0){\includegraphics[width=\unitlength,page=2]{ADM_illustr_v2b.pdf}}%
    \put(0.23030663,0.66217121){\makebox(0,0)[lt]{\lineheight{1.25}\smash{\begin{tabular}[t]{l}Encoder\end{tabular}}}}%
    \put(0.21699566,0.49331207){\makebox(0,0)[lt]{\lineheight{1.25}\smash{\begin{tabular}[t]{l}Predictors\end{tabular}}}}%
    \put(0.70598737,0.49331207){\makebox(0,0)[lt]{\lineheight{1.25}\smash{\begin{tabular}[t]{l}Probes\end{tabular}}}}%
    \put(0,0){\includegraphics[width=\unitlength,page=3]{ADM_illustr_v2b.pdf}}%
  \end{picture}%
\endgroup%

%% file: images/dependence_example_1.pdf_tex
\begingroup%
  \makeatletter%
  \providecommand\color[2][]{%
    \errmessage{(Inkscape) Color is used for the text in Inkscape, but the package 'color.sty' is not loaded}%
    \renewcommand\color[2][]{}%
  }%
  \providecommand\transparent[1]{%
    \errmessage{(Inkscape) Transparency is used (non-zero) for the text in Inkscape, but the package 'transparent.sty' is not loaded}%
    \renewcommand\transparent[1]{}%
  }%
  \providecommand\rotatebox[2]{#2}%
  \newcommand*\fsize{\dimexpr\f@size pt\relax}%
  \newcommand*\lineheight[1]{\fontsize{\fsize}{#1\fsize}\selectfont}%
  \ifx\svgwidth\undefined%
    \setlength{\unitlength}{239.03999329bp}%
    \ifx\svgscale\undefined%
      \relax%
    \else%
      \setlength{\unitlength}{\unitlength * \real{\svgscale}}%
    \fi%
  \else%
    \setlength{\unitlength}{\svgwidth}%
  \fi%
  \global\let\svgwidth\undefined%
  \global\let\svgscale\undefined%
  \makeatother%
  \begin{picture}(1,0.96385542)%
    \lineheight{1}%
    \setlength\tabcolsep{0pt}%
    \put(0,0){\includegraphics[width=\unitlength,page=1]{dependence_example_1.pdf}}%
    \put(0.30581818,0.13169994){\makebox(0,0)[t]{\lineheight{1.25}\smash{\begin{tabular}[t]{c}-1.0\end{tabular}}}}%
    \put(0,0){\includegraphics[width=\unitlength,page=2]{dependence_example_1.pdf}}%
    \put(0.56400001,0.13169994){\makebox(0,0)[t]{\lineheight{1.25}\smash{\begin{tabular}[t]{c}0.0\end{tabular}}}}%
    \put(0,0){\includegraphics[width=\unitlength,page=3]{dependence_example_1.pdf}}%
    \put(0.82218187,0.13169994){\makebox(0,0)[t]{\lineheight{1.25}\smash{\begin{tabular}[t]{c}1.0\end{tabular}}}}%
    \put(0,0){\includegraphics[width=\unitlength,page=4]{dependence_example_1.pdf}}%
    \put(0.53973629,0.07742033){\makebox(0,0)[lt]{\lineheight{1.25}\smash{\begin{tabular}[t]{l}$\rx_1$\end{tabular}}}}%
    \put(0,0){\includegraphics[width=\unitlength,page=5]{dependence_example_1.pdf}}%
    \put(0.2507162,0.20456634){\makebox(0,0)[rt]{\lineheight{1.25}\smash{\begin{tabular}[t]{r}-1.0\end{tabular}}}}%
    \put(0,0){\includegraphics[width=\unitlength,page=6]{dependence_example_1.pdf}}%
    \put(0.2507162,0.48145573){\makebox(0,0)[rt]{\lineheight{1.25}\smash{\begin{tabular}[t]{r}0.0\end{tabular}}}}%
    \put(0,0){\includegraphics[width=\unitlength,page=7]{dependence_example_1.pdf}}%
    \put(0.2507162,0.75834511){\makebox(0,0)[rt]{\lineheight{1.25}\smash{\begin{tabular}[t]{r}1.0\end{tabular}}}}%
    \put(0,0){\includegraphics[width=\unitlength,page=8]{dependence_example_1.pdf}}%
    \put(0.06116241,0.47858098){\makebox(0,0)[lt]{\lineheight{1.25}\smash{\begin{tabular}[t]{l}$\rx_2$\end{tabular}}}}%
    \put(0,0){\includegraphics[width=\unitlength,page=9]{dependence_example_1.pdf}}%
  \end{picture}%
\endgroup%

%% file: images/dependence_example_2.pdf_tex
\begingroup%
  \makeatletter%
  \providecommand\color[2][]{%
    \errmessage{(Inkscape) Color is used for the text in Inkscape, but the package 'color.sty' is not loaded}%
    \renewcommand\color[2][]{}%
  }%
  \providecommand\transparent[1]{%
    \errmessage{(Inkscape) Transparency is used (non-zero) for the text in Inkscape, but the package 'transparent.sty' is not loaded}%
    \renewcommand\transparent[1]{}%
  }%
  \providecommand\rotatebox[2]{#2}%
  \newcommand*\fsize{\dimexpr\f@size pt\relax}%
  \newcommand*\lineheight[1]{\fontsize{\fsize}{#1\fsize}\selectfont}%
  \ifx\svgwidth\undefined%
    \setlength{\unitlength}{239.03999329bp}%
    \ifx\svgscale\undefined%
      \relax%
    \else%
      \setlength{\unitlength}{\unitlength * \real{\svgscale}}%
    \fi%
  \else%
    \setlength{\unitlength}{\svgwidth}%
  \fi%
  \global\let\svgwidth\undefined%
  \global\let\svgscale\undefined%
  \makeatother%
  \begin{picture}(1,0.96385542)%
    \lineheight{1}%
    \setlength\tabcolsep{0pt}%
    \put(0,0){\includegraphics[width=\unitlength,page=1]{dependence_example_2.pdf}}%
    \put(0.30581818,0.13169994){\makebox(0,0)[t]{\lineheight{1.25}\smash{\begin{tabular}[t]{c}-1.0\end{tabular}}}}%
    \put(0,0){\includegraphics[width=\unitlength,page=2]{dependence_example_2.pdf}}%
    \put(0.56400001,0.13169994){\makebox(0,0)[t]{\lineheight{1.25}\smash{\begin{tabular}[t]{c}0.0\end{tabular}}}}%
    \put(0,0){\includegraphics[width=\unitlength,page=3]{dependence_example_2.pdf}}%
    \put(0.82218187,0.13169994){\makebox(0,0)[t]{\lineheight{1.25}\smash{\begin{tabular}[t]{c}1.0\end{tabular}}}}%
    \put(0,0){\includegraphics[width=\unitlength,page=4]{dependence_example_2.pdf}}%
    \put(0.53973629,0.07742033){\makebox(0,0)[lt]{\lineheight{1.25}\smash{\begin{tabular}[t]{l}$\rx_1$\end{tabular}}}}%
    \put(0,0){\includegraphics[width=\unitlength,page=5]{dependence_example_2.pdf}}%
    \put(0.2507162,0.20456634){\makebox(0,0)[rt]{\lineheight{1.25}\smash{\begin{tabular}[t]{r}0.0\end{tabular}}}}%
    \put(0,0){\includegraphics[width=\unitlength,page=6]{dependence_example_2.pdf}}%
    \put(0.2507162,0.48145573){\makebox(0,0)[rt]{\lineheight{1.25}\smash{\begin{tabular}[t]{r}0.5\end{tabular}}}}%
    \put(0,0){\includegraphics[width=\unitlength,page=7]{dependence_example_2.pdf}}%
    \put(0.2507162,0.75834511){\makebox(0,0)[rt]{\lineheight{1.25}\smash{\begin{tabular}[t]{r}1.0\end{tabular}}}}%
    \put(0,0){\includegraphics[width=\unitlength,page=8]{dependence_example_2.pdf}}%
    \put(0.06116241,0.47858098){\makebox(0,0)[lt]{\lineheight{1.25}\smash{\begin{tabular}[t]{l}$\rx_2$\end{tabular}}}}%
    \put(0,0){\includegraphics[width=\unitlength,page=9]{dependence_example_2.pdf}}%
  \end{picture}%
\endgroup%

%% file: images/dependence_example_3.pdf_tex
\begingroup%
  \makeatletter%
  \providecommand\color[2][]{%
    \errmessage{(Inkscape) Color is used for the text in Inkscape, but the package 'color.sty' is not loaded}%
    \renewcommand\color[2][]{}%
  }%
  \providecommand\transparent[1]{%
    \errmessage{(Inkscape) Transparency is used (non-zero) for the text in Inkscape, but the package 'transparent.sty' is not loaded}%
    \renewcommand\transparent[1]{}%
  }%
  \providecommand\rotatebox[2]{#2}%
  \newcommand*\fsize{\dimexpr\f@size pt\relax}%
  \newcommand*\lineheight[1]{\fontsize{\fsize}{#1\fsize}\selectfont}%
  \ifx\svgwidth\undefined%
    \setlength{\unitlength}{239.03999329bp}%
    \ifx\svgscale\undefined%
      \relax%
    \else%
      \setlength{\unitlength}{\unitlength * \real{\svgscale}}%
    \fi%
  \else%
    \setlength{\unitlength}{\svgwidth}%
  \fi%
  \global\let\svgwidth\undefined%
  \global\let\svgscale\undefined%
  \makeatother%
  \begin{picture}(1,0.96385542)%
    \lineheight{1}%
    \setlength\tabcolsep{0pt}%
    \put(0,0){\includegraphics[width=\unitlength,page=1]{dependence_example_3.pdf}}%
    \put(0.30581818,0.13169994){\makebox(0,0)[t]{\lineheight{1.25}\smash{\begin{tabular}[t]{c}-1.0\end{tabular}}}}%
    \put(0,0){\includegraphics[width=\unitlength,page=2]{dependence_example_3.pdf}}%
    \put(0.56400001,0.13169994){\makebox(0,0)[t]{\lineheight{1.25}\smash{\begin{tabular}[t]{c}0.0\end{tabular}}}}%
    \put(0,0){\includegraphics[width=\unitlength,page=3]{dependence_example_3.pdf}}%
    \put(0.82218187,0.13169994){\makebox(0,0)[t]{\lineheight{1.25}\smash{\begin{tabular}[t]{c}1.0\end{tabular}}}}%
    \put(0,0){\includegraphics[width=\unitlength,page=4]{dependence_example_3.pdf}}%
    \put(0.53973629,0.07742033){\makebox(0,0)[lt]{\lineheight{1.25}\smash{\begin{tabular}[t]{l}$\rx_1$\end{tabular}}}}%
    \put(0,0){\includegraphics[width=\unitlength,page=5]{dependence_example_3.pdf}}%
    \put(0.2507162,0.20456634){\makebox(0,0)[rt]{\lineheight{1.25}\smash{\begin{tabular}[t]{r}-1.0\end{tabular}}}}%
    \put(0,0){\includegraphics[width=\unitlength,page=6]{dependence_example_3.pdf}}%
    \put(0.2507162,0.48145573){\makebox(0,0)[rt]{\lineheight{1.25}\smash{\begin{tabular}[t]{r}0.0\end{tabular}}}}%
    \put(0,0){\includegraphics[width=\unitlength,page=7]{dependence_example_3.pdf}}%
    \put(0.2507162,0.75834511){\makebox(0,0)[rt]{\lineheight{1.25}\smash{\begin{tabular}[t]{r}1.0\end{tabular}}}}%
    \put(0,0){\includegraphics[width=\unitlength,page=8]{dependence_example_3.pdf}}%
    \put(0.06116241,0.47858098){\makebox(0,0)[lt]{\lineheight{1.25}\smash{\begin{tabular}[t]{l}$\rx_2$\end{tabular}}}}%
    \put(0,0){\includegraphics[width=\unitlength,page=9]{dependence_example_3.pdf}}%
  \end{picture}%
\endgroup%

%% file: sections/background.tex
We begin by defining statistical independence and correlation measures. 
Let ${\rx_1, \ldots, \rx_d}$ be a finite set of random variables. These variables are mutually independent if and only if their joint cumulative distribution function (CDF) factorizes:
\begin{equation}
F_{\rx_1, \dots, \rx_d}(\evx_1, \dots, \evx_d) = \prod_{i=1}^d F_{\rx_i}(x_i) \quad \forall \evx_1, \dots, \evx_d \in \mathbb{R}
\end{equation}
where $F_{\rx_i}$ denotes the marginal CDF.
Equivalently, mutual independence holds if and only if each variable $\rx_i$ is statistically independent of the vector of all remaining variables $\rvx_{-i}$.
Notably, pairwise independence between all $\rx_i$ and $\rx_j$ does not imply mutual independence~\citep{driscoll1978pairwise_mutual_indep_example}. We refer to Appendix~\ref{subapp:pairwise_vs_mutual_ex} for an example. 

Dependence is often quantified using Pearson's correlation coefficient ($\Cor$), which measures linear relationships.
However, $\Cor(\rx_i, \rx_j) = 0$ does not imply independence, as it fails to capture nonlinear dependencies.
We can instead consider the \textit{maximal correlation coefficient} \cite{gebelein1941mcor_origin,renyi1959measures}, defined as the supremum of Pearson correlations over all square-integrable non-constant transformations $\phi$ and $\psi$:
\begin{equation} \label{eq:max_cor}
    \mCor(\rx_i, \rx_j) \triangleq \sup_{\phi, \psi} \Cor(\psi(\rx_i), \phi(\rx_j)).
\end{equation}
Crucially, $\mCor$ vanishes if and only if the variables are independent \cite{lancaster1960indep_and_zero_cor}. We thus aim to design a method that minimizes the maximal correlation coefficient.

%% file: sections/related_work.tex
\label{sec:related_work}

\begin{table*}[tb]
\centering \small
\caption{Hierarchy of independence guarantees provided by $\Cor(\psi(\cdot),\phi(\cdot)) = 0$ across different method classes. While PCA and Predictability Minimization are limited to linear or mean independence, our \ADM~algorithm provably optimizes for mutual independence.}
\label{tab:mcor_implications_summary}
\begin{tabular}{lllll}
    \toprule
    Method Class & Family of $\psi$ & Family of $\phi$ & Input to $\phi$ & Implication of $\Cor(\psi(\cdot),\phi(\cdot))=0$ \\ \midrule
    Decorrelation & Linear & Linear & $\rx_j \in \R$ & Linear independence \\
    Predictability Min. & Linear & Square-integrable & $\rvx_{-i} \in \R^{d-1}$ & Mean independence \\
    \rowcolor{gray!10} \ADM~(ours) & Square-integrable & Square-integrable & $\rvx_{-i} \in \R^{d-1}$ & Mutual independence \\
    \bottomrule
\end{tabular}
\end{table*}

The Principal Component Analysis (PCA) was the first dimensionality reduction technique focusing on extracting uncorrelated features. 
This transformation is achieved while preserving the maximal variance in the original dataset. 

\paragraph{Autoencoders~(AEs).} 
The PCA reduction is closely related to autoencoders \citep{kramer1991autoencoder_nlpca, rumelhart1986ae_structure}. 
An AE consists of two neural networks: an encoder that compresses the input into a lower-dimensional code, and a decoder that reconstructs the input from this representation. 
Both networks are trained jointly with a reconstruction error. 
Interestingly, the optimal solution of a linear AE corresponds to performing PCA. 
However, unlike PCA, an AE can learn nonlinear dimensionality reductions, but its latent space is not guaranteed to have uncorrelated dimensions. 
The variational autoencoder (VAE) from  \citet{kingma2013VAE} is an extension of AEs in which the encoder maps the input into a probabilistic representation and the decoder reconstructs data by sampling from this distribution. 
This extension of AEs enabled the generation of new samples similar to the training data. 
Multiple works~\citep{higgins2022beta_vae, burgess2018understand_beta_vae, kim2018factor_vae, chen2018tc_vae} proposed variations of the VAE objective to encourage disentangled representations. Intuitively, it implies learning representations where changes in one dimension correspond to changes in a single latent variable, but disentangled concepts may be dependent. 
From this perspective, the problem is ill-defined~\citep{locatello2019challenging_disentanglement} and differs from this work's independence objective. 

\paragraph{Predictability minimization (PM).}
Our framework generalizes the PM principle~\cite{schmidhuber1996predictability_min,schraudolph1999predictability_min}, which learns representations by making individual features unpredictable from the remaining ones. PM uses a minimax objective over reconstruction errors of raw features; at convergence (under standardization), this enforces $\mathbb{E}\left[\rz_i \mid \rvz_{-i}\right] = \mathbb{E}\left[\rz_i\right]$, a condition known as \textit{mean independence}.
While necessary, mean independence is strictly weaker than statistical independence and fails to capture higher-order dependencies (see Appendix \ref{subapp:mean_vs_mutual_indep} for a simple example).
Our method addresses this by transforming targets using \textit{probe networks} and standardizing their outputs to ensure scale invariance; we prove that the resulting minimax equilibrium achieves mutual independence via \textit{maximal correlation} minimization (see Table~\ref{tab:mcor_implications_summary} for a comparison across method classes).
A recent work~\citep{zollikofer2024higher_order_redund_ssl} applies PM to self-supervised learning by predicting randomly masked features via auxiliary networks, but inherits the same limitation as PM, leaving higher-order dependencies unaddressed.

\paragraph{Adversarial learning.} 
We now discuss the core training paradigm behind our algorithm. 
While adversarial principles date back to \textit{Predictability Minimization} \cite{schmidhuber1996predictability_min}, Generative Adversarial Networks (GANs, \cite{goodfellow2014generative}) popularized this framework for high-dimensional synthesis: a generator is trained to generate realistic synthetic data, while a discriminator is trained to predict whether a sample comes from the training dataset or the generator. 
InfoGANs~\cite{chen2016infogan} extends GANs by adding a criterion that maximizes a lower bound of the mutual information (MI) between the representations and the generated data. This helps learn disentangled representations. 
Our approach differs from InfoGANs in that it does not optimize for input-output MI maximization and is not bound to generative networks. 
Most similar to our work, \citet{brakel2017indep_feats_adversarial} used adversarial networks to decrease dependence by training an encoder to produce samples from a joint distribution that are indistinguishable from samples of the product of its marginals. However, this training objective is unstable and requires careful tuning, whereas ours systematically converges. 

\paragraph{Self-supervised learning~(SSL).} 
SSL is an active area of research in representation learning. 
Popularized by SimCLR~\citep{chen2020SimCLR_ssl}, a leading paradigm in SSL is to train a network to be invariant under carefully designed data augmentations. Intuitively, this enforces consistency between the input and the output by pushing two views of the same image to lie close in the embedding space, i.e., a slight change in the input should not lead to a completely different output. 
While effective, this idea comes with the risk of collapsed representations~\citep{hua2021feature_decorr}: dimensional collapse and collapse to a single value. 
The former occurs when information encoded across different dimensions of the representation is redundant~\citep{jing2021understand_ssl_dim_collapse}, whereas the latter is generally due to a lack of a counterweight to the invariance term. 
Many SSL approaches relied on linear decorrelation of output dimensions~\citep{huang2018decorrelated_bn,zbontar2021barlow_ssl,ermolov2021whitening_WMSE_ssl,bardes2021vicreg_ssl} to avoid collapse. 
However, as illustrated in Example~\ref{example:uncorrelated_but_dependent}, linearly uncorrelated dimensions can still co-vary and therefore suffer from dimensional collapse. 
Our algorithm can be used to extend this idea to minimize redundancy in general. 
For a more in-depth discussion of redundancy in SSL, we refer readers to Appendix~\ref{sec:extended_related_works}.

\paragraph{Independent Component Analysis~(ICA).} 
ICA~\citep{jutten1991ica_bss} is a method to separate mixed signals into independent ones, assuming that the observations are a linear mixture of mutually independent, non-Gaussian latent variables. Motivated by the central limit theorem, implementations typically involve the maximization of non-Gaussianity by relying on differential entropy~\citep{hyvarinen2000ICA_book} or mutual information~\citep{bell1995infomax}. When our method’s encoder is a linear layer with equal input and output dimensions, it simplifies to an ICA implementation that does not explicitly optimize for non-Gaussianity. Also, when combined with an AE, our algorithm performs a joint PCA and ICA. However, the successive application of PCA and ICA~\citep{back1997IPCA_first,yao2012ipca_later} is not equivalent to our approach, since PCA may extract redundant features that ICA is then ill-posed on.

\paragraph{Dependence estimation.} Measures of dependence or correlation have been extensively studied across foundational and modern contexts. \citet{renyi1959measures} introduced axioms for dependence measures, proposing maximal correlation as a key metric. Building on this, \citet{hsu2019corresp_nn} used neural networks to model relationships among features via correspondence analysis, whereas our work minimizes rather than merely measures or analyzes dependencies. More recently, \citet{xu2024neural_feat_learning} provided a framework for learning features by encoding statistical dependencies in a structured manner, but without solving a task-specific objective. 
Other notable areas of research that studied measures of correlation include Canonical Correlation Analysis \cite{hotelling1992cca_orig} and its nonlinear extensions \cite{lai2000kcca_1,shotaro2001kcca_2,vinokourov2002kcca_inferring,andrew2013deep_cca}, Alternating Conditional Expectation \cite{breiman1985estimating}, Deep Hirschfield-Gebelein-Rényi \cite{grari2021renyi_min}, etc.

%% file: sections/method.tex
\label{sec:method_description}

We present Adversarial Dependence Minimization (\ADM), an algorithm for learning to extract features that are minimally interdependent.
Consider the representations $\smash{\vz^{(i)} = f_{\theta}(\vx^{(i)})}$ from input samples $\vx^{(i)}$ drawn i.i.d. from a dataset $\smash{\rmX=\{\vx^{(i)}\}_{i=1}^N}$.
Our algorithm involves three types of networks: 
\begin{enumerate}[itemsep=2pt,topsep=0pt,parsep=0pt,partopsep=0pt]
    \item an encoder $f_\theta: \mathcal{X} \to \mathbb{R}^d$ that learns lower-dimensional representations of the training data.
    \item a set of \textit{probe networks} $\vpsi$ that transform each feature $z_i$ into a new variable $\tilde{z}_i = \psi_i(z_i)$ to maximize predictability. Intuitively, probes aim to expose dependencies that are impossible to predict with a simple regression on $z_i$. 
    \item a set of \textit{dependency predictors} $\vphi$ with one predictor for every embedding dimension: $\phi_i: \mathbb{R}^{d-1} \to \mathbb{R}$. 
    The predictors are trained to learn how dimensions relate by 
    estimating the conditional expectation of a (transformed) dimension given the context of all others: $\hat{z}_i = \phi_i(\vz_{-i}) = \phi_i(z_1, \dots, z_{i-1}, z_{i+1}, \dots, z_d)$.
\end{enumerate}

Taking inspiration from GANs~\citep{goodfellow2014generative}, we formulate training as a minimax game: the dependence branch ($\vphi, \vpsi$) serves as a critic, minimizing the reconstruction error to expose dependencies. Simultaneously, the encoder ($\theta$) maximizes the error by driving features toward statistical independence. The objective function is:
\begin{equation} \label{eq:minmax_game}
    \max_{\theta} \min_{\{\vphi, \vpsi\}} \quad \mathbb{E}_{\vz\sim P(\mathcal{X};\theta)} \frac{1}{d} \sum_{i=1}^d \left( \psi_i(z_i) - \phi_i(\vz_{-i}) \right)^2
\end{equation}
where $P(\mathcal{X};\theta)$ denotes the encoder's learned representation distribution, parameterized by $\theta$.

\subsection{Standardization and problem solution} \label{subsec:stdized_formulation}

We may, without loss of generality, assume that probe outputs have zero mean and unit variance. 
This standardization makes the problem scale-invariant and ensures that the objective is well-defined, i.e., has finite and positive variance. Indeed, without constraints, the probe networks could trivially minimize \cref{eq:minmax_game} by collapsing $\psi_i(z_i)$ to a constant (zero variance).

Practically, we standardize the output of the probe networks before computing the loss: $\smash{\tilde{z}_i \leftarrow \left(\tilde{z}_i - \mathbb{E}[\tilde{z}_i]\right)/\sqrt{\mathbb{V}[\tilde{z}_i]} }$. Where the mean $\mathbb{E}[\tilde{z}_i]$ and variance $\mathbb{V}[\tilde{z}_i]$ statistics are estimated over the current mini-batch, similar to Batch Normalization~\citep{ioffe2015batchnorm}.

We proceed by showing that, under standardization, the minimax objective has a global solution corresponding to mutual independence.

\begin{theorem}[Global Optimality] \label{theorem:convergence_std_game}
    Let $\rvz$ be the representation vector. 
    Assume $\rvz$ induces a square-integrable distribution (all functions in $L^2$), and infinite capacity for the predictors $\vphi$ and probes $\vpsi$, and that $\psi_i$ are standardized to have zero mean and unit variance, the minimax solution to \cref{eq:minmax_game} is achieved if and only if the components of $\rvz$ are mutually statistically independent.
\end{theorem}

\begin{proof}[Proof Sketch]
    See Appendix \ref{app:proof_theroem1} for the full proof.
    Briefly, for fixed encoder $\theta$ and probes $\psi$, the optimal predictors $\phi^*$ are the conditional expectations $\mathbb{E}[\psi_j(z_i) | \rvz_{-i}]$.
    Substituting these into the loss yields the sum of conditional variances.
    Minimizing it is equivalent to maximizing the \textit{Explained Variance} (fraction of variance of $\psi_j(z_i)$ explainable by $\rvz_{-i}$).
    By Rényi's theorem~\citep{renyi1959measures}, the supremum of this quantity is exactly the squared Maximal Correlation.
    Thus, the encoder maximizes the loss by forcing $\mCor(\rz_i, \rvz_{-i}) = 0$ for all $i$, which implies mutual independence.
\end{proof}

\begin{corollary} \label{corol:convergence_sol}
    At convergence, dependency predictors predict the constant function $\phi_i\left(\rvz_{-i}\right)=\mathbb{E}[\tilde{\rz}_i]=0$ for all $i$, and the expected average reconstruction error is equal to 1.
\end{corollary}

\subsection{Implementation}

We parameterize both the predictors and probes using Multi-Layer Perceptrons (MLPs), as they are universal function approximators~\citep{hornik_mlp_universal,cybenko1989universal_approx_orig, leshno1993universal_approx_nonpolyn} and can, in theory, approximate arbitrarily well the relation among the variables if given enough capacity. 

\paragraph{Input Standardization.} We further standardize representations before feeding them into the dependence branch to simplify optimization. Although this step is not strictly required, it consistently yields a modest improvement in convergence speed. 

\paragraph{Efficiency via Parallelization.}
Running $d$ separate networks for the predictors and probes naively is computationally expensive. Since all predictors $\phi_i$ share the same architecture but take distinct masked inputs, we implement the full set $\vphi$ as a single batched network: a dense linear layer mapping $\mathbb{R}^d \to \mathbb{R}^{d \cdot h}$ with a precomputed static weight mask zeroing diagonal self-connections, which ensures the $i$-th group of hidden units receives only $\vz_{-i}$ without materializing $d$ separate inputs. Followed by per-feature hidden layers processed jointly via batched \texttt{einsum} operations over weight tensors. 
We apply the same strategy to the probe networks $\vpsi$: rather than looping through $d$ independent MLPs, we stack their parameters and evaluate all probes simultaneously via a single \texttt{einsum} call per layer. Together, these designs significantly reduce computational overhead by exploiting GPUs' parallelization capabilities. 

\paragraph{Preventing Variance Collapse.} Because standardization makes the objective scale-invariant, the pre-standardized representations are free to adopt arbitrary scales. In practice, this can result in variances approaching zero, which may cause numerical instabilities. We therefore apply a hinge loss~\cite{cortes1995svn_margin,tsochantaridis2005large_margin} to the squared standard deviation of the pre-standardized representations, encouraging it to remain above 1.

%% file: sections/applications.tex
\label{sec:applications}
Having developed an algorithm for mutual and nonlinear decorrelation, we explore its applicability across diverse tasks. A key incentive for learning minimally dependent features is that independence implies minimal redundancy \cite{shannon1948mathematical,barlow1961possible}, which may be beneficial in various contexts. 

Consider a task-specific minimization problem with objective function $\mathcal{L}_{\mathrm{task}}$ that we would like to restrict to solutions for which the output dimensions are independent:
\begin{equation} \label{eq:constrained_problem_strict}
    \underset{\theta}{\min} \quad \quad \mathcal{L}_{\mathrm{task}}(\mX, \cdot ; \theta) \quad
    \textrm{s.t.} \quad \rz_1 \indep \rz_2 \indep \cdots \indep \rz_d
\end{equation}
where $\indep$ denotes that each random variable is mutually independent of all others. 

While solving this constrained problem directly is infeasible, the \ADM~algorithm can be seen as a Lagrangian relaxation of the independence constraint.
Indeed, the encoder's objective can be written as a constraint of the form $\mathcal{L}_{\mathrm{adm}}(\cdot) = 0$, which is satisfied when the dimensions are mutually independent.
Therefore, the problem can be relaxed to:
\begin{equation} \label{eq:approx_problem}
    \underset{\theta}{\min} \quad \mathcal{L}_{\mathrm{task}}(\mX, \cdot ; \theta) + \lambda \mathcal{L}_{\mathrm{adm}}(\mX; \theta, \vphi, \vpsi)
\end{equation} 
where $\lambda \in \mathbb{R}$ is a Lagrange multiplier and where the adversarial objective from the encoder is denoted by $\mathcal{L}_{\mathrm{adm}}(\vz;\vphi,\vpsi) = \frac{1}{d}\sum_{i=1}^d 1 - (\psi_i(z_i) - \phi_i(\vz_{-i}))^2$. 

In the following, we investigate three potential applications: an extension of PCA to nonlinear decorrelation (Section~\ref{subsec:pca_application}), the minimization of redundancy in a supervised classification setting (Section~\ref{subsec:classif_application}), and the use of the algorithm to prevent dimensional collapse in SSL (Section~\ref{subsec:ssl_application}). 

\subsection{Principal and Independent Component Analysis} \label{subsec:pca_application}

PCA is a linear dimensionality reduction technique that transforms a set of variables $\smash{\rmX = \{\vx^{(i)}\}_{i=1}^N}$, $\smash{\vx^{(i)} \in \mathbb{R}^l}$ into a set of uncorrelated representations $\smash{\vz^{(i)} = \mW^T\vx^{(i)}}$, $\smash{\vz^{(i)} \in \mathbb{R}^d}$, where the columns of the matrix $\mW \in \mathbb{R}^{l \times d}$ are called principal components. 
Intuitively, PCA identifies directions $\mW_{:,i}$ in the input space along which the variance of the projected data is maximized. Assuming centered data, the PCA reduction solves the following problem:
\begin{equation} \label{eq:pca_max_var_def}
    \max_{\mW} \quad \frac{1}{N-1} \sum_{i=1}^N \lVert \mW^T\vx^{(i)} \rVert_2^2 \quad
    \textrm{s.t.} \quad \mW^T\mW = \mI_d 
\end{equation}
We proceed to discuss two key properties of PCA. 
Firstly, the orthogonality constraint on the projection matrix $\mW$ ensures that the dimensions of the projections are uncorrelated: $\Cov(\rz_i, \rz_j) = 0 \text{ for all } i \neq j$. 
Secondly, PCA can equivalently be defined as finding the projection matrix minimizing the MSE for reconstructions $\hat{\vx}^{(i)} = \mW\vz^{(i)} = \mW\mW^T\vx^{(i)}$. This dual form led to the use of autoencoders to solve PCA~\citep{baldi1989pca_and_sgd,kramer1991autoencoder_nlpca,hinton2006pca_ae_sgd}. 
It is worth noting that the unit-norm constraint on the principal components is not needed in the dual formulation, as the reconstruction error inherently accounts for the scale of the projections. 

Given these observations, we extend the PCA algorithm to non-linear decorrelation by replacing the covariance constraint with our more general dependence minimization objective: 
\begin{equation} \label{eq:extended_pca_loss}
    \min_{\mW} \quad \frac{1}{N} \sum_{i=1}^N \lVert \vx^{(i)} - \mW \vz^{(i)} \rVert_2^2 
    + \lambda \mathcal{L}_{\mathrm{adm}}(\vz^{(i)}; \vphi,\vpsi) 
\end{equation} 
This PCA extension also holds for non-linear AEs, leading to an NLPCA reduction~\citep{kramer1991autoencoder_nlpca} with minimal redundancy. We refer interested readers to Appendix~\ref{subapp:nlpca_extension}. 
In the following, we illustrate the relevance of this PCA extension with a simple example and refer to it as \textit{Principal and Independent Component Analysis} (PICA). 
\begin{example} \label{ex:pca_vs_pica}
    let two independent latent factors $\rv_1$ and $\rv_2$ be uniformly distributed over the interval $[-\sqrt{3},\sqrt{3}]$ (so that $\mathbb{V}[\rv_1] = \mathbb{V}[\rv_2] = 1$) and observations of a random vector defined by: 
    \smash{$
        \rvx = \big[5\rv_1, 3 \cos(2\pi\rv_1/\sqrt{3}), \rv_2 \big]^T
    $}. 
    The observations are linearly uncorrelated by construction: $\Cov[\rx_i,\rx_j] = 0$ for all $i \neq j$, and with descending variances: $\mathbb{V}[\rx_1] = 25$, $\mathbb{V}[\rx_2] = 9/2$, $\mathbb{V}[\rx_3] = 1$. 
    Since PCA extracts uncorrelated variables with maximal variance, its solution for $d=2$ is: $\rvz_{\mathrm{PCA}} = [\rx_1,\rx_2]^T$. Yet, this solution is redundant, since both $\rx_1$ and $\rx_2$ are functions of $\rv_1$. 
    In contrast, the solution to PICA is the maximum variance solution with independent dimensions, namely: $\rvz_{\mathrm{PICA}} = [\rx_1,\rx_3]^T = [5\rv_1,\rv_2]^T$. 
    Unlike PCA, these learned features capture both true latent factors. 
    We further discuss this example and empirically study its solutions in Appendix \ref{subapp:pca_ipca_example_eval}.
\end{example}

\subsection{Classifiers and Generalization} \label{subsec:classif_application}

We observe that a key limitation of naive classifiers is their poor generalization to out-of-distribution (OOD) data. This limitation can be illustrated by the following example.
\begin{example} \label{example:classif_subset}
    Consider a dataset of colored geometric shapes with three training classes: red squares, green triangles, and blue triangles. 
    Because color alone suffices to discriminate these classes, a classifier trained to minimize empirical risk can achieve zero training loss by relying solely on color, without learning shape-related features.
    Consequently, when evaluated on OOD samples such as red triangles, the classifier misclassifies them as red squares.
\end{example} 
\begin{figure}[h!]\vspace{-9px}
    \centering
    \def\svgwidth{0.3\textwidth} 
    {\scriptsize 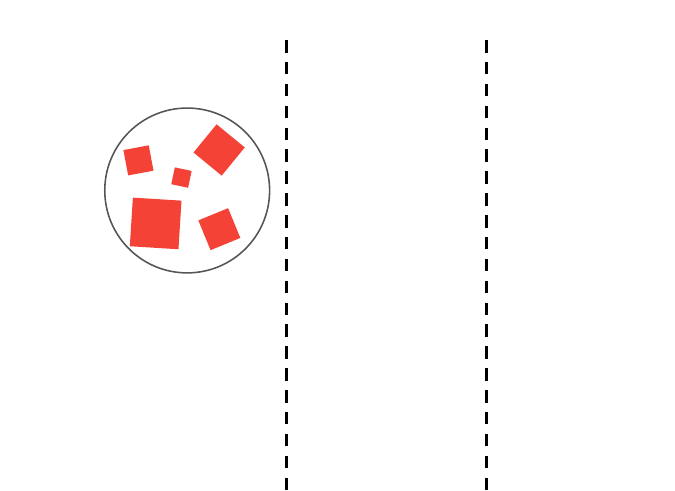}
    \caption{Illustration for Ex. \ref{example:classif_subset}: if a classifier only relied on \textit{color} (dashed lines), it would predict that the red triangle is a red square.}
    \label{fig:example_limit_ce_loss}
\end{figure}

The failure in \cref{example:classif_subset} arises because the classifier ignores the shape.  To mitigate this, we propose using \ADM{} to minimize redundancy among embedding dimensions, encouraging the encoder to capture multiple factors of variation (e.g., shape and color) rather than a single dominant feature. This should help the model generalize to unseen combinations.

Formally, we define the loss function as a weighted combination of a softmax cross-entropy and the \ADM{} loss: 
\begin{multline} \label{eq:cls_min_redund_loss}
    \min_{\theta, W, \vb} \quad \frac{1}{N} \sum_{i=1}^N \mathcal{L}_{\mathrm{CE}}(\softmax(W\vz^{(i)} + \vb), y^{(i)}) 
    \\ + \lambda \mathcal{L}_{\mathrm{adm}}(\vz^{(i)}; \vphi,\vpsi)
\end{multline}
where $\vz^{(i)} \in \mathbb{R}^d$ is the representation generated by the backbone parametrized by $\theta$. The terms $W \in \mathbb{R}^{n_c \times d}$ and $\vb \in \mathbb{R}^{n_c}$ denote the linear head parameters for $n_c$ classes.

\subsection{Self-Supervised Learning} \label{subsec:ssl_application}

SSL aims to learn representations without labels that transfer well to downstream tasks. 
Following common practice~\citep{chen2020SimCLR_ssl,zbontar2021barlow_ssl}, we train a model by pushing two views of the same image to lie close in the embedding space. 
Specifically, given a mini-batch of $n$ images, we duplicate each image, apply different augmentations to the two views, and minimize the MSE between the resulting embeddings $\vz$ and $\vz^\prime$.

While this objective encourages invariance, embeddings can still suffer from full or dimensional collapse (see \cref{sec:related_work}), limiting their usefulness for downstream tasks.
Existing redundancy reduction techniques~\citep{huang2018decorrelated_bn, zbontar2021barlow_ssl,bardes2021vicreg_ssl} mitigate collapse by decorrelating feature dimensions. 
However, linearly uncorrelated embeddings may retain significant redundancy. 
To address this, we use the \ADM{} objective. 
Our approach thus generalizes existing decorrelation techniques. The overall objective is:
\begin{multline} \label{eq:ssl_min_redund_loss}
    \min_\theta \quad \frac{1}{N} \sum_{i=1}^N \lVert \vz^{(i)} - (\vz^\prime)^{(i)} \rVert_2^2 \\
    + \frac{\lambda}{2} \left( \mathcal{L}_{\mathrm{adm}}(\vz^{(i)}; \vphi,\vpsi) + \mathcal{L}_{\mathrm{adm}}((\vz^\prime)^{(i)}; \vphi,\vpsi) \right)
\end{multline}

%% file: images/adversarial_indep_example_1.pdf_tex
\begingroup%
  \makeatletter%
  \providecommand\color[2][]{%
    \errmessage{(Inkscape) Color is used for the text in Inkscape, but the package 'color.sty' is not loaded}%
    \renewcommand\color[2][]{}%
  }%
  \providecommand\transparent[1]{%
    \errmessage{(Inkscape) Transparency is used (non-zero) for the text in Inkscape, but the package 'transparent.sty' is not loaded}%
    \renewcommand\transparent[1]{}%
  }%
  \providecommand\rotatebox[2]{#2}%
  \newcommand*\fsize{\dimexpr\f@size pt\relax}%
  \newcommand*\lineheight[1]{\fontsize{\fsize}{#1\fsize}\selectfont}%
  \ifx\svgwidth\undefined%
    \setlength{\unitlength}{323.73001099bp}%
    \ifx\svgscale\undefined%
      \relax%
    \else%
      \setlength{\unitlength}{\unitlength * \real{\svgscale}}%
    \fi%
  \else%
    \setlength{\unitlength}{\svgwidth}%
  \fi%
  \global\let\svgwidth\undefined%
  \global\let\svgscale\undefined%
  \makeatother%
  \begin{picture}(1,0.72664717)%
    \lineheight{1}%
    \setlength\tabcolsep{0pt}%
    \put(0.52599387,0.69173381){\color[rgb]{0,0,0}\makebox(0,0)[lt]{\lineheight{1.25}\smash{\begin{tabular}[t]{l}Color\end{tabular}}}}%
    \put(0.25530534,0.63119726){\color[rgb]{0,0,0}\makebox(0,0)[lt]{\lineheight{1.25}\smash{\begin{tabular}[t]{l}red\end{tabular}}}}%
    \put(0.53549252,0.63027056){\color[rgb]{0,0,0}\makebox(0,0)[lt]{\lineheight{1.25}\smash{\begin{tabular}[t]{l}green\end{tabular}}}}%
    \put(0.84014454,0.62980721){\color[rgb]{0,0,0}\makebox(0,0)[lt]{\lineheight{1.25}\smash{\begin{tabular}[t]{l}blue\end{tabular}}}}%
    \put(0,0){\includegraphics[width=\unitlength,page=1]{adversarial_indep_example_1.pdf}}%
    \put(0.2080669,0.55564819){\color[rgb]{0.33333333,0.33333333,0.33333333}\rotatebox{28.66}{\makebox(0,0)[lt]{\lineheight{1.25}\smash{\begin{tabular}[t]{l}{\tiny c}\end{tabular}}}}}%
    \put(0.22136502,0.5627606){\color[rgb]{0.33333333,0.33333333,0.33333333}\rotatebox{23.63}{\makebox(0,0)[lt]{\lineheight{1.25}\smash{\begin{tabular}[t]{l}{\tiny l}\end{tabular}}}}}%
    \put(0.22829209,0.56602721){\color[rgb]{0.33333333,0.33333333,0.33333333}\rotatebox{18.34}{\makebox(0,0)[lt]{\lineheight{1.25}\smash{\begin{tabular}[t]{l}{\tiny a}\end{tabular}}}}}%
    \put(0.24355944,0.57105455){\color[rgb]{0.33333333,0.33333333,0.33333333}\rotatebox{11.88}{\makebox(0,0)[lt]{\lineheight{1.25}\smash{\begin{tabular}[t]{l}{\tiny s}\end{tabular}}}}}%
    \put(0.25646371,0.57378831){\color[rgb]{0.33333333,0.33333333,0.33333333}\rotatebox{6.07}{\makebox(0,0)[lt]{\lineheight{1.25}\smash{\begin{tabular}[t]{l}{\tiny s}\end{tabular}}}}}%
    \put(0.27645722,0.57554903){\color[rgb]{0.33333333,0.33333333,0.33333333}\rotatebox{-3.54}{\makebox(0,0)[lt]{\lineheight{1.25}\smash{\begin{tabular}[t]{l}{\tiny 1}\end{tabular}}}}}%
    \put(0,0){\includegraphics[width=\unitlength,page=2]{adversarial_indep_example_1.pdf}}%
    \put(0.50600035,0.25750623){\color[rgb]{0.33333333,0.33333333,0.33333333}\rotatebox{28.66}{\makebox(0,0)[lt]{\lineheight{1.25}\smash{\begin{tabular}[t]{l}{\tiny c}\end{tabular}}}}}%
    \put(0.51929847,0.26461864){\color[rgb]{0.33333333,0.33333333,0.33333333}\rotatebox{23.63}{\makebox(0,0)[lt]{\lineheight{1.25}\smash{\begin{tabular}[t]{l}{\tiny l}\end{tabular}}}}}%
    \put(0.52622554,0.26788525){\color[rgb]{0.33333333,0.33333333,0.33333333}\rotatebox{18.34}{\makebox(0,0)[lt]{\lineheight{1.25}\smash{\begin{tabular}[t]{l}{\tiny a}\end{tabular}}}}}%
    \put(0.54149289,0.27288942){\color[rgb]{0.33333333,0.33333333,0.33333333}\rotatebox{11.88}{\makebox(0,0)[lt]{\lineheight{1.25}\smash{\begin{tabular}[t]{l}{\tiny s}\end{tabular}}}}}%
    \put(0.55439716,0.27564635){\color[rgb]{0.33333333,0.33333333,0.33333333}\rotatebox{6.07}{\makebox(0,0)[lt]{\lineheight{1.25}\smash{\begin{tabular}[t]{l}{\tiny s}\end{tabular}}}}}%
    \put(0.57439068,0.27740707){\color[rgb]{0.33333333,0.33333333,0.33333333}\rotatebox{-3.54}{\makebox(0,0)[lt]{\lineheight{1.25}\smash{\begin{tabular}[t]{l}{\tiny 2}\end{tabular}}}}}%
    \put(0,0){\includegraphics[width=\unitlength,page=3]{adversarial_indep_example_1.pdf}}%
    \put(0.25379945,0.11259381){\color[rgb]{1,1,1}\makebox(0,0)[lt]{\lineheight{1.25}\smash{\begin{tabular}[t]{l}{\large ?}\end{tabular}}}}%
    \put(0,0){\includegraphics[width=\unitlength,page=4]{adversarial_indep_example_1.pdf}}%
    \put(0.80075987,0.25878044){\color[rgb]{0.33333333,0.33333333,0.33333333}\rotatebox{28.66}{\makebox(0,0)[lt]{\lineheight{1.25}\smash{\begin{tabular}[t]{l}{\tiny c}\end{tabular}}}}}%
    \put(0.81405798,0.26589285){\color[rgb]{0.33333333,0.33333333,0.33333333}\rotatebox{23.63}{\makebox(0,0)[lt]{\lineheight{1.25}\smash{\begin{tabular}[t]{l}{\tiny l}\end{tabular}}}}}%
    \put(0.82098505,0.26915946){\color[rgb]{0.33333333,0.33333333,0.33333333}\rotatebox{18.34}{\makebox(0,0)[lt]{\lineheight{1.25}\smash{\begin{tabular}[t]{l}{\tiny a}\end{tabular}}}}}%
    \put(0.8362524,0.2741868){\color[rgb]{0.33333333,0.33333333,0.33333333}\rotatebox{11.88}{\makebox(0,0)[lt]{\lineheight{1.25}\smash{\begin{tabular}[t]{l}{\tiny s}\end{tabular}}}}}%
    \put(0.84915668,0.27692056){\color[rgb]{0.33333333,0.33333333,0.33333333}\rotatebox{6.07}{\makebox(0,0)[lt]{\lineheight{1.25}\smash{\begin{tabular}[t]{l}{\tiny s}\end{tabular}}}}}%
    \put(0.86915019,0.27868128){\color[rgb]{0.33333333,0.33333333,0.33333333}\rotatebox{-3.54}{\makebox(0,0)[lt]{\lineheight{1.25}\smash{\begin{tabular}[t]{l}{\tiny 3}\end{tabular}}}}}%
    \put(0.03491335,0.24946713){\color[rgb]{0,0,0}\rotatebox{90}{\makebox(0,0)[lt]{\lineheight{1.25}\smash{\begin{tabular}[t]{l}Shape\end{tabular}}}}}%
    \put(0.09229914,0.39943005){\color[rgb]{0,0,0}\rotatebox{90}{\makebox(0,0)[lt]{\lineheight{1.25}\smash{\begin{tabular}[t]{l}square\end{tabular}}}}}%
    \put(0.09229914,0.09609858){\color[rgb]{0,0,0}\rotatebox{90}{\makebox(0,0)[lt]{\lineheight{1.25}\smash{\begin{tabular}[t]{l}triangle\end{tabular}}}}}%
  \end{picture}%
\endgroup%

%% file: sections/experiments.tex
Our experimental evaluation aims to assess the algorithm's convergence and study the potential of the supervised and SSL applications.

\begin{figure*} 
    \centering  
    \begin{subfigure}{0.33\textwidth}
        \def\svgwidth{\textwidth}
        {\scriptsize 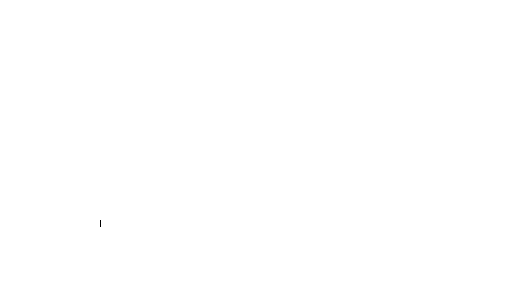}
    \end{subfigure}
    \begin{subfigure}{0.33\textwidth}
        \def\svgwidth{\textwidth}
        {\scriptsize 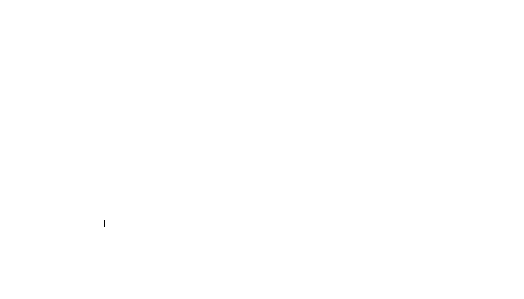}
    \end{subfigure}
    \vspace{-4px}
    \caption{Convergence analysis on TinyImageNet. \textbf{Left:} the reconstruction error of \ADM{} converges to 1. \textbf{Right:} logarithmic plot of the distance correlation on the validation set: $\dCor$ decreases over time, and \ADM{} reaches much lower levels than the linear decorrelation loss.}
    \label{fig:TIN_convergence} 
\end{figure*}

\subsection{Convergence} \label{sec:results_convergence}

This section analyzes the convergence of the adversarial game. 
Specifically, we trained a ResNet-18~\citep{he2016resnet} on the TinyImageNet dataset~\citep{le2015tinyimagenet} without data augmentations. 
We compared \ADM{} to a linear decorrelation baseline.
Detailed experimental setups are provided in Appendix~\ref{subapp:setup_convergence}.

\paragraph{Metrics.} We estimated nonlinear dependencies using the distance correlation \citep{szekely2007_dcorr} between one dimension and the random vector formed by the remaining $d-1$ dimensions. We reported the average of the estimates across the first 64 dimensions. Distance correlation is a non-negative coefficient that characterizes correlations between random vectors by measuring the distance between their joint characteristic function and the product of the marginal characteristic functions. For more details, see Appendix \ref{subapp:dcorr_details}. Of particular interest, the distance correlation is zero $\mathrm{d} \Cor\left(\rx_1, \rx_2\right)=0$ if and only if $\rx_1$ and $\rx_2$ are independent.

\paragraph{Results.} The MSE and distance correlation for the TinyImageNet experiment are reported in \Cref{fig:TIN_convergence}. 
While both methods reduce correlation, adversarial dependence minimization reached a significantly lower final distance correlation than the linear decorrelation approach. 
Furthermore, \ADM{} converged to an average reconstruction error of 1 as expected by Corollary~\ref{corol:convergence_sol}. 
These results support the convergence hypothesis and empirically demonstrate the benefits of \ADM{} over linear decorrelation. 

\subsection{Redundancy Minimization} \label{sec:results_infomax}

This section investigates the impact of our algorithm on redundancy reduction in a classification setting. 
To facilitate the analysis of the representations, the classifiers were trained on a synthetic dataset with known latent factors. 

\paragraph{Dataset.} The Clevr-4~\citep{vaze2024clevr4} dataset is an extension of the CLEVR dataset~\citep{johnson2017clevr}. 
It comprises 100,000 synthetic images of 3D objects of various shapes, colors, textures, and counts. Each taxonomy has 10 classes. 
The label for one taxonomy is sampled uniformly and independently from the others, which means that knowing the label for one taxonomy provides no information about the other taxonomies. 

\paragraph{Evaluation protocols.} We investigated generalization capabilities by training a classifier on one taxonomy (\textit{shape}) and evaluating its accuracy on the remaining taxonomies to assess whether representations encode features beyond those relevant to the training classes. The model is compared with a baseline classifier trained without the adversarial objective. 
We evaluate accuracy using a simple weighted nearest-neighbor (kNN) classifier trained on top of frozen features~\citep{wu2018unsupervised, caron2021DINO_ssl}. 
The kNN algorithm classifies points based on the majority class of their nearest neighbors in the embedding space, providing an easy way to assess clustering quality for each taxonomy. 

\paragraph{Implementation details.} The supervised models are trained for 200 epochs with two data augmentations: random horizontal flipping and cropping. 
We train ResNet-18 encoders. For \ADM{}, we used two-layer dependency predictors and probes. The networks are trained alternately, with one step for each network per iteration. More details are provided in Appendix~\ref{subapp:setup_clevr4}. 

\paragraph{Results.}
We compared our regularized classifier to the baseline classifier. Table~\ref{tab:results_clevr4_infomax} reports the validation accuracy and feature correlation. 
The embedding dimensions of the baseline are highly correlated, with an average squared distance correlation of 0.409. Furthermore, performance on the taxonomies for which the model received no supervision is low, as expected, since it was not incentivized to retain information about them. 
When combining the cross-entropy loss with our adversarial objective, distance correlation drops to 0.038 and the accuracy on the \textit{texture} and \textit{color} taxonomies rises significantly. 
These results validate that \ADM{} reduces redundancy, leading to representations that generalize better. 

\begin{table}[ht]
    \small
    \centering
    \caption{kNN evaluation of the classification approaches on the Clevr-4 dataset. Both are trained with a cross-entropy (c-e) loss. \ADM{} significantly reduces redundancy and helps the classifier generalize beyond the training taxonomy (shape).}
    \label{tab:results_clevr4_infomax}
    \begin{tabular}{l rrrr r} 
    \toprule
        \multirow{2}{*}{method} & \multicolumn{4}{c}{kNN top-1 accuracy [\%]} & \multirow{2}{*}{$\dCor^2$}  \\ 
         & shape & texture & color & count & \\ 
    \midrule
        c-e                    & \textbf{100.0} & 25.0 & 16.4 & \textbf{36.1} &  0.409 \\ 
        \rowcolor{lightgray}
        c-e + \ADM{}  & \textbf{99.9} & \textbf{81.8} & \textbf{98.9} & \textbf{36.0} & \textbf{0.038}  \\ 
    \bottomrule
    \end{tabular}
\end{table}

\subsection{Self-Supervised Learning}
\label{subsec:imagenet_ssl}

This section examines whether our SSL approach prevents dimensional collapse and compares its performance to common approaches on ImageNet. 

\paragraph{Dataset.} ImageNet-1k~\citep{deng2009imagenet} is a standard benchmark for SSL. It contains 1,281,167 training images across 1000 classes. This dataset provides a diverse collection of real-world images. 

\paragraph{Implementation and evaluation.} We trained a ResNet-50 backbone with a three-layer projection head with output dimension 512, and two-layer dependency predictors. 
At evaluation, the projection head is discarded, and the quality of the representations is evaluated by training a linear classifier on top of the backbone with its weights frozen. 
The detailed experimental setup is described in Appendix~\ref{subapp:setup_imagenet_ssl}. 

\paragraph{Results and discussion.} The main SSL methods are compared in \Cref{tab:results_imagenet_ssl}. The average squared distance correlation achieved by our method is 0.022, indicating that the learned representation largely avoids dimensional collapse~\citep{jing2021understand_ssl_dim_collapse}. This value is also substantially lower than those obtained by prior decorrelation-based approaches, such as Barlow Twins and VICReg, highlighting the benefit of accounting for nonlinear dependencies rather than only linear correlations. 
Nevertheless, the method does not match the performance of the strongest SSL baselines. 
A potential explanation is that, while eliminating inter-dimensional redundancy maximizes representational capacity for a fixed latent dimensionality, these representations may not be organized in a form readily exploitable by standard evaluation protocols: kNN and linear probing favor embeddings whose semantic structure is directly reflected in their geometry. Consequently, classification from such representations may require more expressive decoders. In summary, \ADM{} is suboptimal for SSL as it may make features only nonlinearly extractable.

\begin{table}[ht!]
    \small
    \centering
    \caption{Linear evaluation of SSL techniques trained with a ResNet-50 backbone on the ImageNet-1k dataset. The average squared distance correlation is estimated on the output representation.}
    \label{tab:results_imagenet_ssl}
    \begin{tabular}{lrr}
    \toprule
        method  & acc. [\%] & $\smash{\dCor^2}$ \\
    \midrule
        MoCo \citep{he2020MoCo_ssl}                 & 60.6 & 0.111 \\
        SimCLR \citep{chen2020SimCLR_ssl}           & 69.3 & --- \\
        Barlow Twins \citep{zbontar2021barlow_ssl}  & 73.2 & 0.128 \\
        VICReg \citep{bardes2021vicreg_ssl}         & 73.2 & 0.130 \\
        BYOL \citep{grill2020BYOL_ssl}              & 74.3 & 0.145 \\ 
        DINO \citep{caron2021DINO_ssl}              & \textbf{75.3} & 0.490 \\
        \rowcolor{lightgray}
        ours                                        & 65.3 & \textbf{0.022} \\
    \bottomrule
    \end{tabular}
    \vskip -0.1in
\end{table}

%% file: images/TIN_convergence_loss.pdf_tex
\begingroup%
  \makeatletter%
  \providecommand\color[2][]{%
    \errmessage{(Inkscape) Color is used for the text in Inkscape, but the package 'color.sty' is not loaded}%
    \renewcommand\color[2][]{}%
  }%
  \providecommand\transparent[1]{%
    \errmessage{(Inkscape) Transparency is used (non-zero) for the text in Inkscape, but the package 'transparent.sty' is not loaded}%
    \renewcommand\transparent[1]{}%
  }%
  \providecommand\rotatebox[2]{#2}%
  \newcommand*\fsize{\dimexpr\f@size pt\relax}%
  \newcommand*\lineheight[1]{\fontsize{\fsize}{#1\fsize}\selectfont}%
  \ifx\svgwidth\undefined%
    \setlength{\unitlength}{251.775884bp}%
    \ifx\svgscale\undefined%
      \relax%
    \else%
      \setlength{\unitlength}{\unitlength * \real{\svgscale}}%
    \fi%
  \else%
    \setlength{\unitlength}{\svgwidth}%
  \fi%
  \global\let\svgwidth\undefined%
  \global\let\svgscale\undefined%
  \makeatother%
  \begin{picture}(1,0.5687449)%
    \lineheight{1}%
    \setlength\tabcolsep{0pt}%
    \put(0,0){\includegraphics[width=\unitlength,page=1]{TIN_convergence_loss.pdf}}%
    \put(0.19180043,0.09118354){\makebox(0,0)[t]{\lineheight{1.25}\smash{\begin{tabular}[t]{c}0\end{tabular}}}}%
    \put(0,0){\includegraphics[width=\unitlength,page=2]{TIN_convergence_loss.pdf}}%
    \put(0.37090706,0.09118354){\makebox(0,0)[t]{\lineheight{1.25}\smash{\begin{tabular}[t]{c}10000\end{tabular}}}}%
    \put(0,0){\includegraphics[width=\unitlength,page=3]{TIN_convergence_loss.pdf}}%
    \put(0.55001367,0.09118354){\makebox(0,0)[t]{\lineheight{1.25}\smash{\begin{tabular}[t]{c}20000\end{tabular}}}}%
    \put(0,0){\includegraphics[width=\unitlength,page=4]{TIN_convergence_loss.pdf}}%
    \put(0.72912029,0.09118354){\makebox(0,0)[t]{\lineheight{1.25}\smash{\begin{tabular}[t]{c}30000\end{tabular}}}}%
    \put(0,0){\includegraphics[width=\unitlength,page=5]{TIN_convergence_loss.pdf}}%
    \put(0.90822691,0.09118354){\makebox(0,0)[t]{\lineheight{1.25}\smash{\begin{tabular}[t]{c}40000\end{tabular}}}}%
    \put(0.55001367,0.03685693){\makebox(0,0)[t]{\lineheight{1.25}\smash{\begin{tabular}[t]{c}Training step\end{tabular}}}}%
    \put(0,0){\includegraphics[width=\unitlength,page=6]{TIN_convergence_loss.pdf}}%
    \put(0.14608726,0.17225129){\makebox(0,0)[rt]{\lineheight{1.25}\smash{\begin{tabular}[t]{r}0.2\end{tabular}}}}%
    \put(0,0){\includegraphics[width=\unitlength,page=7]{TIN_convergence_loss.pdf}}%
    \put(0.14608726,0.2392583){\makebox(0,0)[rt]{\lineheight{1.25}\smash{\begin{tabular}[t]{r}0.4\end{tabular}}}}%
    \put(0,0){\includegraphics[width=\unitlength,page=8]{TIN_convergence_loss.pdf}}%
    \put(0.14608726,0.30626535){\makebox(0,0)[rt]{\lineheight{1.25}\smash{\begin{tabular}[t]{r}0.6\end{tabular}}}}%
    \put(0,0){\includegraphics[width=\unitlength,page=9]{TIN_convergence_loss.pdf}}%
    \put(0.14608726,0.37327236){\makebox(0,0)[rt]{\lineheight{1.25}\smash{\begin{tabular}[t]{r}0.8\end{tabular}}}}%
    \put(0,0){\includegraphics[width=\unitlength,page=10]{TIN_convergence_loss.pdf}}%
    \put(0.14608726,0.44027939){\makebox(0,0)[rt]{\lineheight{1.25}\smash{\begin{tabular}[t]{r}1.0\end{tabular}}}}%
    \put(0,0){\includegraphics[width=\unitlength,page=11]{TIN_convergence_loss.pdf}}%
    \put(0.14608726,0.50728642){\makebox(0,0)[rt]{\lineheight{1.25}\smash{\begin{tabular}[t]{r}1.2\end{tabular}}}}%
    \put(0.05877623,0.34465672){\rotatebox{90}{\makebox(0,0)[t]{\lineheight{1.25}\smash{\begin{tabular}[t]{c}Mean squared error\end{tabular}}}}}%
    \put(0,0){\includegraphics[width=\unitlength,page=12]{TIN_convergence_loss.pdf}}%
    \put(0.79837019,0.19317156){\makebox(0,0)[lt]{\lineheight{1.25}\smash{\begin{tabular}[t]{l}ADM\end{tabular}}}}%
  \end{picture}%
\endgroup%

%% file: images/TIN_convergence_dcor.pdf_tex
\begingroup%
  \makeatletter%
  \providecommand\color[2][]{%
    \errmessage{(Inkscape) Color is used for the text in Inkscape, but the package 'color.sty' is not loaded}%
    \renewcommand\color[2][]{}%
  }%
  \providecommand\transparent[1]{%
    \errmessage{(Inkscape) Transparency is used (non-zero) for the text in Inkscape, but the package 'transparent.sty' is not loaded}%
    \renewcommand\transparent[1]{}%
  }%
  \providecommand\rotatebox[2]{#2}%
  \newcommand*\fsize{\dimexpr\f@size pt\relax}%
  \newcommand*\lineheight[1]{\fontsize{\fsize}{#1\fsize}\selectfont}%
  \ifx\svgwidth\undefined%
    \setlength{\unitlength}{251.5025bp}%
    \ifx\svgscale\undefined%
      \relax%
    \else%
      \setlength{\unitlength}{\unitlength * \real{\svgscale}}%
    \fi%
  \else%
    \setlength{\unitlength}{\svgwidth}%
  \fi%
  \global\let\svgwidth\undefined%
  \global\let\svgscale\undefined%
  \makeatother%
  \begin{picture}(1,0.56936313)%
    \lineheight{1}%
    \setlength\tabcolsep{0pt}%
    \put(0,0){\includegraphics[width=\unitlength,page=1]{TIN_convergence_dcor.pdf}}%
    \put(0.19937674,0.09128266){\makebox(0,0)[t]{\lineheight{1.25}\smash{\begin{tabular}[t]{c}0\end{tabular}}}}%
    \put(0,0){\includegraphics[width=\unitlength,page=2]{TIN_convergence_dcor.pdf}}%
    \put(0.34618642,0.09128266){\makebox(0,0)[t]{\lineheight{1.25}\smash{\begin{tabular}[t]{c}20\end{tabular}}}}%
    \put(0,0){\includegraphics[width=\unitlength,page=3]{TIN_convergence_dcor.pdf}}%
    \put(0.49299611,0.09128266){\makebox(0,0)[t]{\lineheight{1.25}\smash{\begin{tabular}[t]{c}40\end{tabular}}}}%
    \put(0,0){\includegraphics[width=\unitlength,page=4]{TIN_convergence_dcor.pdf}}%
    \put(0.63980577,0.09128266){\makebox(0,0)[t]{\lineheight{1.25}\smash{\begin{tabular}[t]{c}60\end{tabular}}}}%
    \put(0,0){\includegraphics[width=\unitlength,page=5]{TIN_convergence_dcor.pdf}}%
    \put(0.78661546,0.09128266){\makebox(0,0)[t]{\lineheight{1.25}\smash{\begin{tabular}[t]{c}80\end{tabular}}}}%
    \put(0,0){\includegraphics[width=\unitlength,page=6]{TIN_convergence_dcor.pdf}}%
    \put(0.93342509,0.09128266){\makebox(0,0)[t]{\lineheight{1.25}\smash{\begin{tabular}[t]{c}100\end{tabular}}}}%
    \put(0.56640092,0.03689699){\makebox(0,0)[t]{\lineheight{1.25}\smash{\begin{tabular}[t]{c}Epoch\end{tabular}}}}%
    \put(0,0){\includegraphics[width=\unitlength,page=7]{TIN_convergence_dcor.pdf}}%
    \put(0.17154402,0.15931024){\makebox(0,0)[rt]{\lineheight{1.25}\smash{\begin{tabular}[t]{r}0.02\end{tabular}}}}%
    \put(0,0){\includegraphics[width=\unitlength,page=8]{TIN_convergence_dcor.pdf}}%
    \put(0.17154402,0.23724314){\makebox(0,0)[rt]{\lineheight{1.25}\smash{\begin{tabular}[t]{r}0.04\end{tabular}}}}%
    \put(0,0){\includegraphics[width=\unitlength,page=9]{TIN_convergence_dcor.pdf}}%
    \put(0.17154402,0.31517602){\makebox(0,0)[rt]{\lineheight{1.25}\smash{\begin{tabular}[t]{r}0.08\end{tabular}}}}%
    \put(0,0){\includegraphics[width=\unitlength,page=10]{TIN_convergence_dcor.pdf}}%
    \put(0.17154402,0.39310891){\makebox(0,0)[rt]{\lineheight{1.25}\smash{\begin{tabular}[t]{r}0.16\end{tabular}}}}%
    \put(0,0){\includegraphics[width=\unitlength,page=11]{TIN_convergence_dcor.pdf}}%
    \put(0.17154402,0.4710418){\makebox(0,0)[rt]{\lineheight{1.25}\smash{\begin{tabular}[t]{r}0.32\end{tabular}}}}%
    \put(0.05884012,0.34503136){\rotatebox{90}{\makebox(0,0)[t]{\lineheight{1.25}\smash{\begin{tabular}[t]{c}$\dCor$\end{tabular}}}}}%
    \put(0,0){\includegraphics[width=\unitlength,page=12]{TIN_convergence_dcor.pdf}}%
    \put(0.6346023,0.47473808){\makebox(0,0)[lt]{\lineheight{1.25}\smash{\begin{tabular}[t]{l}decorrelation\end{tabular}}}}%
    \put(0,0){\includegraphics[width=\unitlength,page=13]{TIN_convergence_dcor.pdf}}%
    \put(0.6346023,0.41637633){\makebox(0,0)[lt]{\lineheight{1.25}\smash{\begin{tabular}[t]{l}ADM\end{tabular}}}}%
  \end{picture}%
\endgroup%

%% file: sections/forward.tex
\label{sec:forward}

This paper laid the foundations for learning mutually independent embedding dimensions, but several important questions and applications remain to be studied. 

\paragraph{Compression-accessibility tradeoff.} As illustrated with the self-supervised learning application, learning highly compressed, information-dense representations can limit the efficacy of simple downstream heads. Because eliminating all inter-dimensional dependencies scatters the semantic structure across a warped manifold, using these features for downstream tasks such as classification will likely require deeper, non-linear decoders. Optimizing purely for compression may thus come at the cost of feature accessibility. Studying this tradeoff is an interesting open question. 

\paragraph{Controllable capacity.} \ADM{} currently relies on a fixed, pre-defined number of embedding dimensions. 
This may, however, not always be ideal. Consider, for instance, the case where more embedding dimensions are available than latent factors; the method may split a single concept across several dimensions (for instance, encoding an integer in $\{1,\ldots,16\}$ across four binary dimensions $\{0,1\}^4$). 
One potential extension would be to adapt \ADM{} to have a controllable capacity, for instance by optimizing for using a minimal number of dimensions, similar to the idea in~\citet{burgess2018understand_beta_vae}.

\paragraph{Applications.} The mechanism from \ADM{} may benefit several application domains, including:
\begin{itemize}[itemsep=2pt,topsep=0pt,parsep=0pt,partopsep=0pt]
    \item \emph{Open-set recognition:} This task requires a classifier to accurately categorize instances from known training classes while identifying novel, unseen classes during inference \cite{scheirer2012toward_osr}. While existing approaches primarily focus on defining compact decision boundaries, success also depends on representations retaining information beyond class-specific cues. Our classification results (\Cref{sec:results_infomax}) suggest that reducing redundancy promotes richer, more transferable representations, potentially supporting compositional generalization and improving the detection of unknown classes.
    \item \emph{Multi-modal learning:} In vision-language models, \ADM~could be applied to separate shared and modality-specific components across modalities, improving alignment and fusion by explicitly structuring the latent space into complementary subspaces.
    \item \emph{Domain adaptation:} Similarly, the independence objective provides a mechanism to separate domain-invariant from domain-specific information into two sets of features, with the potential to make domain shifts more explicit in the representation space.
\end{itemize}
Note that in multi-modal learning and domain adaptation, \ADM~would operate across sets of embeddings rather than across embedding dimensions.

%% file: sections/conclusion.tex
This work studied the problem of learning representations whose dimensions are statistically independent and introduced the first general algorithm for obtaining such features. We showed that the minimax optimum of the proposed objective is attained if and only if the learned dimensions are mutually independent, providing a principled foundation for representation learning with independent dimensions.

We further developed a computationally efficient implementation that scales to high-dimensional embeddings with thousands of dimensions. 
Using this implementation, we showed that \ADM{} can recover principal and independent components, that reducing redundancy in supervised classification can improve generalization beyond the training classes, and that it prevents dimensional collapse in SSL, albeit at the cost of less accessible features. 

While many questions remain open (see \Cref{sec:forward}), our results suggest that representational independence is a promising direction for representation learning: explicitly reducing redundancy may improve generalization, promote more compositional representations, and enable more effective compression.

%% file: appendix/proof_theroem1.tex
\section{Proofs} 

\subsection{Proof of Theorem 1} \label{app:proof_theroem1}
The proof proceeds by solving the optimization problem in stages: first for the predictors $\vphi$, second for the probes $\vpsi$, and finally for the encoder $\theta$.

\paragraph{Optimization of the Predictors.} 
Fix the encoder $\theta$ (and thus the distribution of embeddings $\rvz$) and the probes $\vpsi$.
The value function is a sum of non-negative terms, so we can minimize each term independently with respect to the corresponding $\phi_i$.

Consider the i-th MSE term: $\min_{\phi_i} \mathbb{E} \left[ (\psi_i(\rz_i) - \phi_i(\rvz_{- i}))^2 \right]$. We seek for the min-cost function $\phi_i^*\left(\rvz_{- i}\right)$. By the fundamental property of conditional expectation in $L^2$ spaces \cite{williams1991probability}, the function that minimizes the $L^2$ distance to a target random variable $y=\psi_i\left(\rz_i\right)$ is the orthogonal projection of $y$ onto the subspace of functions measurable w.r.t. $\rvz_{- i}$. This projection is exactly the conditional expectation:
\begin{equation}
    \phi_i^*(\rvz_{- i}) = \mathbb{E}[\psi_i(\rz_i) \mid \rvz_{- i}]
\end{equation}

\paragraph{Value at the optimum.} 
By the definition of conditional variance:
\begin{equation}
\mathbb{V}[\psi_i(\rz_i) \mid \rvz_{- i}]=\mathbb{E}\left[(\psi_i(\rz_i) - \mathbb{E}[\psi_i(\rz_i) \mid \rvz_{- i}])^2 \mid \rvz_{- i}\right]
\end{equation}

If we take the expectation of the conditional variance over $\rvz_{- i}$ and use the Law of Total Expectation, we get:
\begin{align}
\mathbb{E}[\mathbb{V}[\psi_i(\rz_i) \mid \rvz_{- i}]] &= \mathbb{E}\left[\mathbb{E}\left[(\psi_i(\rz_i)-\mathbb{E}[\psi_i(\rz_i) \mid \rvz_{- i}])^2 \mid \rvz_{- i}\right]\right] \\
&= \mathbb{E}\left[(\psi_i(\rz_i)-\mathbb{E}[\psi_i(\rz_i) \mid \rvz_{- i}])^2\right]
\end{align}

We now substitute $\phi_i^*$ into the objective. Given that $\phi_i^*$ is the conditional mean, the MSE is thus the expected conditional variance, i.e., the \textit{unexplained variance}:
\begin{align}
\min _{\phi_i} \mathcal{L}_i &= \mathbb{E} \left[ (\psi_i(\rz_i) - \mathbb{E}[\psi_i(\rz_i) \mid \rvz_{-i}])^2 \right] \\
&= \mathbb{E}\left[\mathbb{V}\left[\psi_i\left(\rz_i\right) \mid \rvz_{- i} \right] \right]
\end{align}

Finally, using the Law of Total Variance and imposing the standardization constraint $\mathbb{V}\left[\psi_i\left(\rz_i\right)\right]=1$, we find:
\begin{align}
    \mathbb{E}\left[\mathbb{V}\left[\psi_i\left(\rz_i\right) \mid \rvz_{- i} \right] \right] &= \mathbb{V}\left[\psi_i\left(\rz_i\right)\right]-\mathbb{V}\left[\mathbb{E}\left[\psi_i\left(\rz_i\right) \mid \rvz_{- i}\right]\right] \\
    &= 1-\mathbb{V}\left[\mathbb{E}\left[\psi_i\left(\rz_i\right) \mid \rvz_{- i}\right]\right]
\end{align}

\paragraph{Optimization of the Probes.}
The adversaries $(\vphi, \vpsi)$ cooperate to minimize the value function. Minimizing the loss with respect to $\psi_i$ is equivalent to maximizing the subtracted term:
\begin{equation}
    \min_{\psi_i} \left( 1 - \mathbb{V}\left[(\mathbb{E}[\psi_i(\rz_i) \mid \rvz_{- i}]\right] \right) \iff \max_{\psi_i} \mathbb{V}\left[\mathbb{E}[\psi_i(\rz_i) \mid \rvz_{- i}]\right]
\end{equation}
subject to $\mathbb{E}[\psi_i(\rz_i)] = 0$ and $\mathbb{V}[\psi_i(\rz_i)] = 1$.

The square of the Maximal Correlation Coefficient \cite{gebelein1941mcor_origin,renyi1959measures} between a variable $\ru$ and a vector $\rvv$, denoted $\mCor^2(\ru, \rvv)$, is defined as:
\begin{equation}
\mCor^2(\ru, \rvv)=\sup_{\phi, \psi}(\Cor(\psi(\ru), \phi(\rvv)))^2=\sup _{\psi} \mathbb{V}[\mathbb{E}[\psi(\ru) \mid \rvv]] \quad \text{s.t.} \quad \mathbb{V}[\psi(\ru)]=1
\end{equation}
where the supremum is taken over all zero-mean, unit-variance $L^2$ functions.

Therefore, the inner optimization for each dimension $i$ computes exactly the maximal correlation between $\rz_i$ and the rest of the vector $\rvz_{- i}$ :
\begin{equation}
\max _{\psi_i} \mathbb{V}\left[\mathbb{E}\left[\psi_i\left(\rz_i\right) \mid \rvz_{- i}\right]\right]=\mCor^2\left(\rz_i, \rvz_{- i}\right)
\end{equation}
The value of the game after the full inner optimization is: 
\begin{equation}
    \min_{\vpsi, \vphi} \mathcal{L} = \frac{1}{d} \sum_{i=1}^d (1 - \mCor_i^2(\rz_i,\rvz_{-i}))
\end{equation}

\paragraph{Optimization of the Encoder.}

The outer loop maximizes the encoder parameters $\theta$:
\begin{equation}
\max_\theta \frac{1}{d} \sum_{i=1}^d\left(1-\mCor_i^2\left(\rz_i,\rvz_{-i}\right)\right) \Longleftrightarrow \min_\theta \sum_{i=1}^d \mCor_i^2\left(\rz_i,\rvz_{-i}\right)
\end{equation}

Since $\mCor^2$ is a squared correlation, we have $0 \leq \mCor^2 \leq 1$. The sum is minimized (and the original objective maximized to 1) if and only if:
\begin{equation}
\forall i \in\{1, \ldots, d\}, \quad \mCor_i^2\left(\rz_i,\rvz_{-i}\right)=0
\end{equation}

A fundamental property of the Maximal Correlation Coefficient \cite{renyi1959measures} is that it vanishes if and only if the variables are independent.
\begin{equation}
\mCor_i^2\left(\rz_i,\rvz_{-i}\right)=0 \Longleftrightarrow \rz_i \perp \rvz_{-i}
\end{equation}

Indeed, if $\mCor^2=0$, then $\mathbb{V}\left[\mathbb{E}\left[\psi_i\left(\rz_i\right) \mid \rvz_{- i}\right]\right]=0$ for all valid $\psi_i$. Since expectations are zero, this implies $\mathbb{E}\left[\psi_i\left(\rz_i\right) \mid \rvz_{-i}\right]=0$ almost surely for all square-integrable $\psi_i$. By the definition of conditional independence, this implies the distribution of $\rz_i$ does not depend on $\rvz_{-i}$.

Since this holds for all $i$, each coordinate $\rz_i$ is independent of the rest. Iteratively applying this factorization to the joint density yields $p(\vz)=\prod_i p(z_i)$, i.e., mutual independence of all dimensions.

\hfill $\square$

%% file: appendix/pca_extension.tex
\section{Principal and Independent Component Analysis (PICA)}

\begin{figure}
    \begin{subfigure}{\textwidth}
        \centering
        \includegraphics[width=0.66\linewidth]{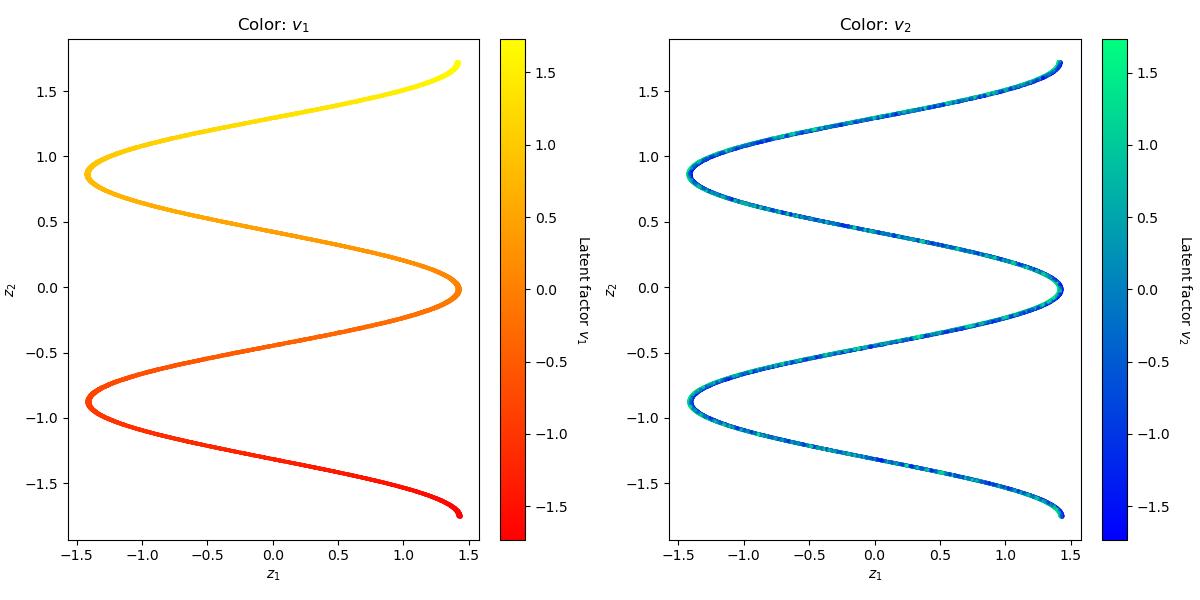}
        \caption{PCA reduction implemented with a linear autoencoder and a covariance regularization term.}
        \label{fig:latent_space_pca_covreg}
    \end{subfigure}
    \vspace{20px}
    \begin{subfigure}{\textwidth}
        \centering
        \includegraphics[width=0.66\linewidth]{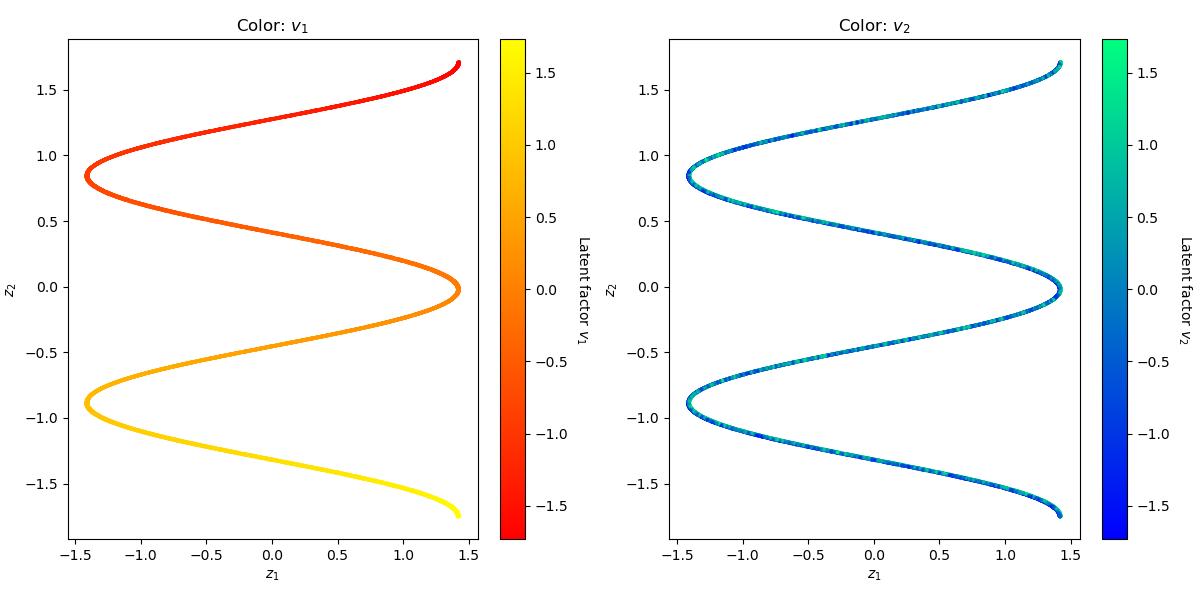}
        \caption{PCA reduction implemented with a linear autoencoder and linear \textit{dependency predictors}.}
        \label{fig:latent_space_pca_lin_dependency}
    \end{subfigure}
    \vspace{20px}
    \begin{subfigure}{\textwidth}
        \centering
        \includegraphics[width=0.66\linewidth]{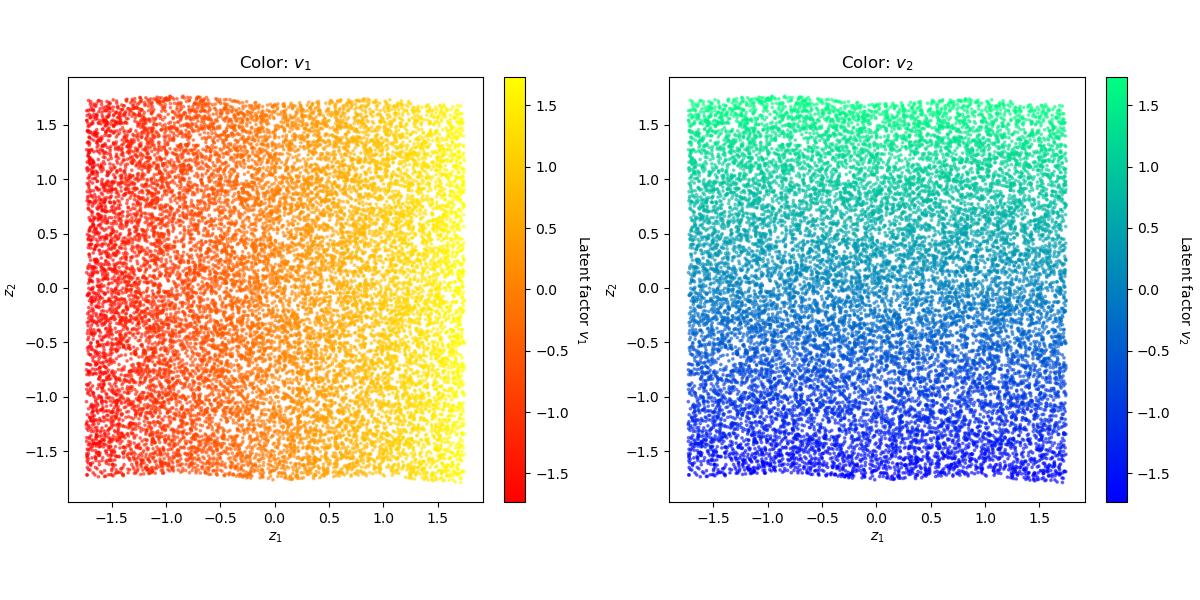}
        \caption{PICA reduction implemented with a linear autoencoder and nonlinear \textit{dependency predictors}.}
        \label{fig:latent_space_pica_nonlin_dependency}
    \end{subfigure}
    \caption{Learned representations $\rvz$. The colors indicate the value of the original latent factors $\rv_1$ (left) and $\rv_2$ (right). }
    \label{fig:latent_spaces_pca_pica_example}
\end{figure}

\subsection{Empirical Study of Example~\ref{ex:pca_vs_pica}} \label{subapp:pca_ipca_example_eval}

In Example~\ref{ex:pca_vs_pica}, the solution that maximizes the explained variance under the zero correlation constraint is $\rvz_{\mathrm{PCA}} = [\rx_1,\rx_2]^T$, with a total variance of $\mathbb{V}[\rvz_{\mathrm{PCA}}] = 25 + 4.5 = 29.5$. 
However, this is not a solution to PICA since $\rx_1 = 5 \rv_1$ and $\rx_2 = 3 \cos{2\pi\rv_1/\sqrt{3}}$ are both functions of the same latent factor $\rv_1$. 
The solution to PICA is thus $\rvz_{\mathrm{PICA}} = [\rx_1,\rx_3]^T$, with a total explained variance of $\mathbb{V}[\rvz_{\mathrm{PICA}}] = 25 + 1 = 26$. 
Attention should be drawn to the fact that the explained variance of PICA is always smaller than or equal to the PCA decomposition. Equality occurs only when the highest-variance uncorrelated combination of the inputs is mutually independent. 

We empirically study this example by comparing the solutions to four different implementations:
\begin{enumerate}
    \item the PCA decomposition solved with a Singular Value Decomposition. For this example, we rely on the \textit{scikit-learn}~\citep{scikit-learn} implementation: \url{https://scikit-learn.org/1.6/modules/generated/sklearn.decomposition.PCA.html}.
    \item the PCA decomposition implemented with a linear autoencoder and a covariance minimization objective. This approach is similar to~\cite{mialon2022vcreg} but without their variance regularization term. 
    \item the PCA decomposition implemented with a linear autoencoder and our standardized adversarial objective with linear dependency predictors. 
    \item the PICA decomposition implemented with a linear autoencoder and our standardized adversarial objective with nonlinear dependency predictors and probe networks. 
\end{enumerate}

\paragraph{Implementation details.} 
We train methods (2) to (4) for 5000 steps. We generate 512 observations $\rmX$ by sampling from the uniform latent factors $\rvv$ at every iteration. 
The encoder is implemented with a projection $\mW^T \in \mathbb{R}^{3 \times 2}$ and the decoder uses the same matrix $\mW$. The autoencoders and dependence branch are trained using the Adam optimizer, with learning rates of $5 \cdot 10^{-3}$ and $2 \cdot 10^{-2}$, respectively. The training step ratio is set to $k=16$. 
The covariance/dependence minimization loss coefficients are set to $\lambda=1$, while the reconstruction loss coefficient is set to 0.02. This weighting strategy aims to prevent the method from compromising covariance/dependence in pursuit of a decreased reconstruction error.  

The PCA implementations from (1), (2), and (3) find, respectively:
\begin{align}
    W_{\mathrm{PCA,1}} \approx 
    \begin{bmatrix}
        1 & 0 & 0 \\
        0 & 1 & 0
    \end{bmatrix} 
    \quad W_{\mathrm{PCA,2}} \approx 
    \begin{bmatrix}
        0.00 & 1.00 & -0.01 \\
        1.00 & -0.01 & 0.00
    \end{bmatrix} 
    \quad W_{\mathrm{PCA,3}} \approx 
    \begin{bmatrix}
        0.00 & 1.00 & 0.01 \\
        -1.00 & 0.00 & 0.01
    \end{bmatrix} 
\end{align} 
In line with theory, the three PCA implementations extracted the first two observed variables $\rx_1$ and $\rx_2$. 
For implementations (1) to (3), the total explained variances for the representations $\rz$ are all approximately equal to $29.5$, and the covariance matrices are close to diagonal. 
However, the learned latent dimensions are not independent, with a distance correlation of $\dCor(\rz_1, \rz_2) \approx 0.25$.
The methods accurately reconstruct the first two observed variables $\rx_1$ and $\rx_2$, and achieve an average reconstruction error of 1 for $\rx_3$. It should be emphasized that this average error corresponds to the variance of $\rx_3$. 

The representations $\rvz$ for experiments (2) and (3) are shown in Figure~\ref{fig:latent_space_pca_covreg} and Figure~\ref{fig:latent_space_pca_lin_dependency}. The abscissa and ordinate depict the learned latent variables $\rz_1$ and $\rz_2$, while the color of the predictions indicates the true value of the true latent factors $\rv_1$ (left figure) and $\rv_2$ (right figure). The distributions align with our previous finding and illustrate that none of the PCA implementations encoded the latent factor $\rv_2$. 

On the contrary, the PICA implementation (4) finds:
\begin{align}
    W_{\mathrm{PICA}} \approx 
    \begin{bmatrix}
        1.00 & 0.00 & 0.01 \\
        0.00 & -0.01 & 1.00
    \end{bmatrix} 
\end{align}
The average reconstruction error is 4.46, corresponding to the variance of the second dimension. 
Furthermore, the distance correlation between the two learned representations is $\dCor(\rz_1, \rz_2) \approx 0.008$. 
These empirical results validate that the PICA reduction captured the two independent latent factors $\rv_1$ and $\rv_2$. Furthermore, the method could not reconstruct $\rx_2$ because it uses a linear autoencoder. 
The distribution of the learned representations $\rvz$ is shown in Figure~\ref{fig:latent_space_pica_nonlin_dependency}. 

\subsection{Nonlinear Principal and Independent Component Analysis} \label{subapp:nlpca_extension}

The Non-Linear Principal Component Analysis (NLPCA) extended PCA to non-linear transformations. 
It was introduced by \citet{kramer1991autoencoder_nlpca} as an autoencoder that learns the identity mapping. The architecture included an intermediate "bottleneck" representation, forcing the network to learn low-dimensional data representations. 
The original architecture was a four-layer feed-forward neural network with sigmoid activation functions. 
Following this definition, one can learn an NLPCA reduction by training an encoder $f_{\vtheta}: \mathbb{R}^l \to \mathbb{R}^d$ and a decoder $g_{\xi}: \mathbb{R}^d \to \mathbb{R}^l$ to minimize the reconstruction error: 
\begin{equation} \label{eq:original_nlpca}
    \min_{\vtheta,\xi} \quad \quad \frac{1}{N} \sum_{i=1}^N \lVert \vx^{(i)} - g_{\xi}(f_{\vtheta}(\vx^{(i)})) \rVert_2^2
\end{equation}
One may assume that a solution with minimal reconstruction error should exhibit low redundancy in the bottleneck representation. 
Yet, \cref{eq:original_nlpca} does not explicitly push the bottleneck representation to have uncorrelated dimensions. 
Therefore, starting from the encoder-decoder architecture from the \cref{eq:original_nlpca}, we add the \ADM{} objective by following the development from \cref{subsec:pca_application}:
\begin{equation}
    \min_{\vtheta,\xi} \quad \quad \frac{1}{N} \sum_{i=1}^N \lVert \vx^{(i)} - g_{\xi}(f_{\vtheta}(\vx^{(i)})) \rVert_2^2 
    + \lambda \mathcal{L}_{\mathrm{adm}}(\vz^{(i)}; \vphi, \vpsi)
\end{equation}
where $\vz^{(i)} = f_{\vtheta}(\vx^{(i)})$ is the bottleneck representation. 
We denote this extension as \textit{Non-Linear Principal and Independent Component Analysis} (NLPICA). It extends the NLPCA reduction by adding an objective that pushes the representations to have minimally interdependent dimensions. 

Note that the solution to the NLPICA reduction is not unique. 
In fact, research on nonlinear independent component analysis (NLICA) has demonstrated that there are numerous ways to transform independent variables while preserving statistical independence~\citep{darmois1951analyse_liaisons,jutten2004_BSS_NLICA}. These transformations can involve complex mixing functions that yield representations that may be challenging to interpret or exhibit undesirable properties in certain applications. 
For examples of mixing transformations, see~\citep{taleb1999example_mixing}. 
Despite this limitation, the problem can be made identifiable again with additional assumptions such as temporal non-stationarity~\citep{hyvarinen2016time_contrast_ica}, a conditionally factorized prior distribution over the latent variables of a VAE~\citep{khemakhem2020ivae_ica} or constraining the function class with constraints on their partial derivatives~\citep{buchholz2022ident_nlica_fn_class}.

%% file: appendix/extended_related.tex
\section{Extended Related Work}
\label{sec:extended_related_works}

\paragraph{SSL and optimality of uncorrelatedness.} Several studies in SSL highlighted the importance of uncorrelatedness. Specifically, \citet{johnson2022contrast_optim_casis} showed that contrastive learning approximates a positive-pair kernel, and applying Kernel PCA to this kernel yields orthogonal representations that are optimal for linear fine-tuning. More broadly, \citet{zhai2023understand_ssl_rkhs} demonstrated that the optimal d-dimensional encoder aligns with the top-d eigenspace of the augmentation-induced kernel operator. 
These findings suggest that optimal representations should have orthogonal dimensions --- a property our approach explicitly optimizes for, as independence implies uncorrelatedness.
\citet{li2021ssl_kernel_HSIC} built upon this idea and extended BYOL \cite{grill2020BYOL_ssl} with the Hilbert-Schmidt Independence Criterion (HSIC), a kernel-based measure of dependence between probability distributions. However, unlike our method, this method's decorrelation effect depends on the choice of kernel. 
Finally, \citet{shwartz2024compress} discussed SSL from an information-theoretic standpoint, highlighting the importance of both compression and invariance through the multi-view assumption \citep{sridharan2008information_multiview}. 

\paragraph{Input-output mutual information maximization.} 
Another line of work on dimensionality reduction is based on the infomax principle~\citep{linsker1988infomax,bell1995infomax}, which aims to maximize the mutual information (MI) between the input data and the neural network's output to learn informative representations. 
\citet{belghazi2018MINE} introduced Mutual Information Neural Estimation~(MINE), a first estimate of the MI between high-dimensional continuous random variables using neural networks. Based on MINE, DeepInfoMax~\citep{hjelm2019deep_infomax} learns representations by maximizing the input-output MI, maximizing the MI between global and local representations, and matching the output to a uniform prior with adversarial learning. 
Similarly to our work's objective, MI is an information-theoretic measure of the information shared by random variables. However, unlike those methods, our algorithm minimizes dependencies among output dimensions rather than maximizing a proxy for the input-output MI. Furthermore, our algorithm is not limited to pairwise dependencies.

%% file: appendix/independence.tex
\section{Independence and Correlation} 

\subsection{Pairwise Independence without Mutual Independence} \label{subapp:pairwise_vs_mutual_ex}

\begin{figure}[h!]
    \centering
    \begin{subfigure}{0.49\textwidth}
        \centering
        \includegraphics[width=0.7\linewidth]{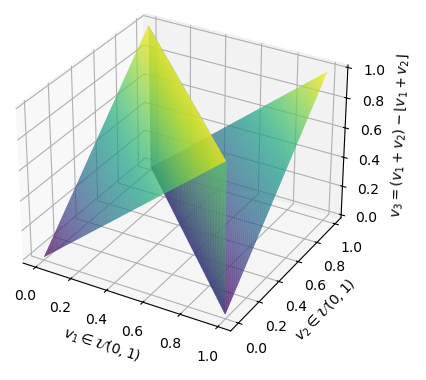}
    \end{subfigure}
    \hfill
    \begin{subfigure}{0.49\textwidth}
        \centering
        \includegraphics[width=0.74\linewidth]{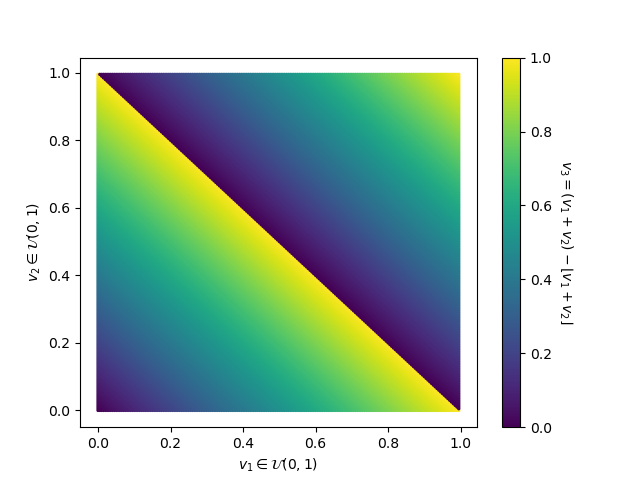}
    \end{subfigure}
    \caption{Illustration of \Cref{example:pairwise_vs_mutual}: the random variables $\rx_1$, $\rx_2$ and $\rx_3$ are all mutually dependent despite being all pairwise independent.}
    \label{fig:pairwise_not_mutual_illustrs}
\end{figure}

\begin{example} \label{example:pairwise_vs_mutual}
    Let two random variables $\rx_1$ and $\rx_2$ be drawn from uniform distributions in the interval $\left[0,1\right]$. Then, define the random variable $\rx_3 = \rx_1 + \rx_2 - \floor{\rx_1 + \rx_2}$ where $\floor{\cdot}$ denotes the integral part of the value (e.g., $\floor{3.14} = 3$).
    \newline The random variable $\rx_3$ is uniformly distributed on $\left[0, 1\right]$ (see Figure~\ref{fig:pairwise_not_mutual_illustrs}) and the pairs $\{\rx_1, \rx_2\}$, $\{\rx_1, \rx_3\}$ and $\{\rx_2, \rx_3\}$ are pairwise independent. Still, the variables are not mutually independent since the uncertainty of $\rx_3$ is zero when the random vector $\rvx_{-3} = [\rx_1, \rx_2]^T$ is known.
\end{example}

\subsection{Mean Independence without Full Independence}
\label{subapp:mean_vs_mutual_indep}

In this section, we provide a concrete example of a continuous bivariate distribution where $\rx_2$ is mean-independent of $\rx_1$ but not fully independent. This example illustrates a scenario where predictability minimization \cite{schmidhuber1996predictability_min} would not succeed, even if our standardized formulation were adopted. 

\begin{example} \label{example:mean_vs_full_indep}
    Let $\rx_1$ be a continuous random variable uniformly distributed on the interval $[1, 2]$. 
    We define the conditional distribution of $\rx_2$ given $\rx_1$ as a normal distribution centered at zero with a variance scaling quadratically with $\rx_1$:
    \begin{equation}
        \rx_2 \mid \rx_1 \sim \mathcal{N}(0, \rx_1^2).
    \end{equation}
    
    To establish mean independence, we verify that the conditional expectation $\mathbb{E}[\rx_2 \mid \rx_1]$ equals the marginal expectation $\mathbb{E}[\rx_2]$. By definition, the conditional mean is invariant to $\rx_1$: $\mathbb{E}[\rx_2 \mid \rx_1] = 0$.
    
    Applying the law of total expectation yields the marginal expectation:
    \begin{equation}
        \mathbb{E}[\rx_2] = \mathbb{E}_{\rx_1} [ \mathbb{E}_{\rx_2\mid \rx_1}[\rx_2 \mid \rx_1] ] = \mathbb{E}_{\rx_1}[0] = 0.
    \end{equation}
    Since $\mathbb{E}[\rx_2 \mid \rx_1] = \mathbb{E}[\rx_2] = 0$, $\rx_2$ is mean-independent of $\rx_1$.
    
    However, $\rx_1$ and $\rx_2$ are not fully independent. Full independence requires all conditional moments to be invariant to the conditioning variable. The conditional variance of $\rx_2$ given $\rx_1$ is:
    \begin{equation}
        \mathbb{V}[\rx_2 \mid \rx_1] = \rx_1^2.
    \end{equation}
    
    Because the conditional variance depends on the value of $\rx_1$, the joint density function does not factorize into the product of the marginals. Consequently, full statistical independence does not hold.
\end{example}

\begin{figure}[h!] 
    \centering
    \def\svgwidth{0.4\textwidth}
    {\scriptsize 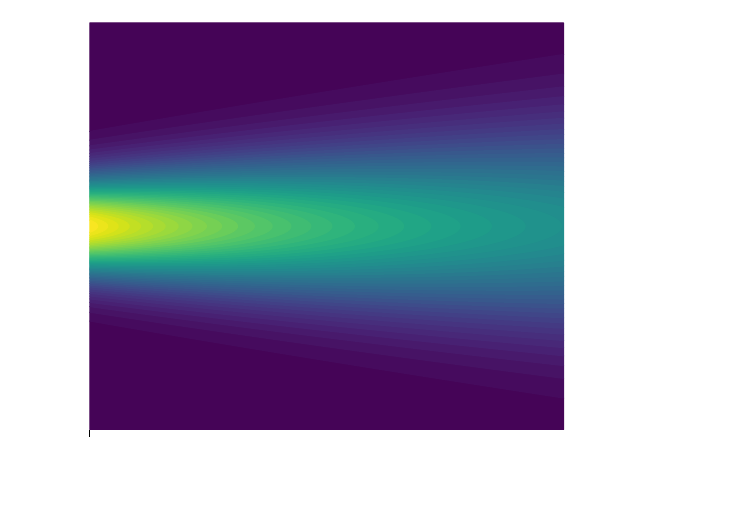}
    \caption{Illustration of \Cref{example:mean_vs_full_indep}: joint density $f(x_1, x_2)$ with mean independence without full statistical independence due to heteroscedastic variance.}
    \label{fig:mean_indep_example}
\end{figure} 

\subsection{Distance Correlation} \label{subapp:dcorr_details}

Distance correlation~\citep{szekely2007_dcorr} is a non-negative coefficient that characterizes both linear and nonlinear correlations between random vectors. 
Let $\rx_1$ and $\rx_2$ be two random vectors with finite first moments, their respective characteristic functions be denoted $\psi_{\rx_1}$ and $\psi_{\rx_2}$, and their joint characteristic function be denoted $\psi_{\rx_1,\rx_2}$. Distance covariance measures the distance between their joint characteristic function and the product of the marginal characteristic functions:
\begin{equation}
    \mathcal{V}^2(\rx_1, \rx_2)=\int_{\mathbb{R}^{p+q}}\left|\psi_{\rx_1, \rx_2}(t, s)-\psi_{\rx_1}(t) \psi_{\rx_2}(s)\right|^2 w(t, s) d t d s
\end{equation}
where $w(t, s)$ is a positive weight function and characteristic functions are $\psi_{\rx}(t)=\mathbb{E}\left[e^{i t \rx}\right]$. Analogous to Pearson correlation, the squared distance correlation $\dCor^2$ is defined by: $\dCor^2(\rx_1,\rx_2) = \mathcal{V}^2(\rx_1,\rx_2) / \sqrt{\mathcal{V}^2(\rx_1, \rx_1) \mathcal{V}^2(\rx_2, \rx_2)}$ if $\mathcal{V}^2(\rx_1, \rx_1) \mathcal{V}^2(\rx_2, \rx_2)>0$ and $0$ otherwise. 

We refer to the original paper~\citep{szekely2007_dcorr} for the empirical estimation of distance correlation and to the library~\citep{carreno2023_dcor_python} for its implementation.

%% file: images/mean_indep_example.pdf_tex
\begingroup%
  \makeatletter%
  \providecommand\color[2][]{%
    \errmessage{(Inkscape) Color is used for the text in Inkscape, but the package 'color.sty' is not loaded}%
    \renewcommand\color[2][]{}%
  }%
  \providecommand\transparent[1]{%
    \errmessage{(Inkscape) Transparency is used (non-zero) for the text in Inkscape, but the package 'transparent.sty' is not loaded}%
    \renewcommand\transparent[1]{}%
  }%
  \providecommand\rotatebox[2]{#2}%
  \newcommand*\fsize{\dimexpr\f@size pt\relax}%
  \newcommand*\lineheight[1]{\fontsize{\fsize}{#1\fsize}\selectfont}%
  \ifx\svgwidth\undefined%
    \setlength{\unitlength}{351.257862bp}%
    \ifx\svgscale\undefined%
      \relax%
    \else%
      \setlength{\unitlength}{\unitlength * \real{\svgscale}}%
    \fi%
  \else%
    \setlength{\unitlength}{\svgwidth}%
  \fi%
  \global\let\svgwidth\undefined%
  \global\let\svgscale\undefined%
  \makeatother%
  \begin{picture}(1,0.69468152)%
    \lineheight{1}%
    \setlength\tabcolsep{0pt}%
    \put(0,0){\includegraphics[width=\unitlength,page=1]{mean_indep_example.pdf}}%
    \put(0.12212805,0.06615064){\makebox(0,0)[t]{\lineheight{1.25}\smash{\begin{tabular}[t]{c}1.0\end{tabular}}}}%
    \put(0,0){\includegraphics[width=\unitlength,page=2]{mean_indep_example.pdf}}%
    \put(0.25186373,0.06615064){\makebox(0,0)[t]{\lineheight{1.25}\smash{\begin{tabular}[t]{c}1.2\end{tabular}}}}%
    \put(0,0){\includegraphics[width=\unitlength,page=3]{mean_indep_example.pdf}}%
    \put(0.38159942,0.06615064){\makebox(0,0)[t]{\lineheight{1.25}\smash{\begin{tabular}[t]{c}1.4\end{tabular}}}}%
    \put(0,0){\includegraphics[width=\unitlength,page=4]{mean_indep_example.pdf}}%
    \put(0.5113351,0.06615064){\makebox(0,0)[t]{\lineheight{1.25}\smash{\begin{tabular}[t]{c}1.6\end{tabular}}}}%
    \put(0,0){\includegraphics[width=\unitlength,page=5]{mean_indep_example.pdf}}%
    \put(0.64107078,0.06615064){\makebox(0,0)[t]{\lineheight{1.25}\smash{\begin{tabular}[t]{c}1.8\end{tabular}}}}%
    \put(0,0){\includegraphics[width=\unitlength,page=6]{mean_indep_example.pdf}}%
    \put(0.77080647,0.06615064){\makebox(0,0)[t]{\lineheight{1.25}\smash{\begin{tabular}[t]{c}2.0\end{tabular}}}}%
    \put(0.44646726,0.02721025){\makebox(0,0)[t]{\lineheight{1.25}\smash{\begin{tabular}[t]{c}$\rx_1$\end{tabular}}}}%
    \put(0,0){\includegraphics[width=\unitlength,page=7]{mean_indep_example.pdf}}%
    \put(0.10219967,0.09689507){\makebox(0,0)[rt]{\lineheight{1.25}\smash{\begin{tabular}[t]{r}-6\end{tabular}}}}%
    \put(0,0){\includegraphics[width=\unitlength,page=8]{mean_indep_example.pdf}}%
    \put(0.10219967,0.18950449){\makebox(0,0)[rt]{\lineheight{1.25}\smash{\begin{tabular}[t]{r}-4\end{tabular}}}}%
    \put(0,0){\includegraphics[width=\unitlength,page=9]{mean_indep_example.pdf}}%
    \put(0.10219967,0.28211395){\makebox(0,0)[rt]{\lineheight{1.25}\smash{\begin{tabular}[t]{r}-2\end{tabular}}}}%
    \put(0,0){\includegraphics[width=\unitlength,page=10]{mean_indep_example.pdf}}%
    \put(0.10219967,0.37472338){\makebox(0,0)[rt]{\lineheight{1.25}\smash{\begin{tabular}[t]{r}0\end{tabular}}}}%
    \put(0,0){\includegraphics[width=\unitlength,page=11]{mean_indep_example.pdf}}%
    \put(0.10219967,0.4673328){\makebox(0,0)[rt]{\lineheight{1.25}\smash{\begin{tabular}[t]{r}2\end{tabular}}}}%
    \put(0,0){\includegraphics[width=\unitlength,page=12]{mean_indep_example.pdf}}%
    \put(0.10219967,0.55994225){\makebox(0,0)[rt]{\lineheight{1.25}\smash{\begin{tabular}[t]{r}4\end{tabular}}}}%
    \put(0,0){\includegraphics[width=\unitlength,page=13]{mean_indep_example.pdf}}%
    \put(0.10219967,0.65255168){\makebox(0,0)[rt]{\lineheight{1.25}\smash{\begin{tabular}[t]{r}6\end{tabular}}}}%
    \put(0.04212984,0.38553941){\rotatebox{90}{\makebox(0,0)[t]{\lineheight{1.25}\smash{\begin{tabular}[t]{c}$\rx_2$\end{tabular}}}}}%
    \put(0,0){\includegraphics[width=\unitlength,page=14]{mean_indep_example.pdf}}%
    \put(0.85906011,0.09689507){\makebox(0,0)[lt]{\lineheight{1.25}\smash{\begin{tabular}[t]{l}0.000\end{tabular}}}}%
    \put(0,0){\includegraphics[width=\unitlength,page=15]{mean_indep_example.pdf}}%
    \put(0.85906011,0.16357386){\makebox(0,0)[lt]{\lineheight{1.25}\smash{\begin{tabular}[t]{l}0.048\end{tabular}}}}%
    \put(0,0){\includegraphics[width=\unitlength,page=16]{mean_indep_example.pdf}}%
    \put(0.85906011,0.23025264){\makebox(0,0)[lt]{\lineheight{1.25}\smash{\begin{tabular}[t]{l}0.096\end{tabular}}}}%
    \put(0,0){\includegraphics[width=\unitlength,page=17]{mean_indep_example.pdf}}%
    \put(0.85906011,0.29693147){\makebox(0,0)[lt]{\lineheight{1.25}\smash{\begin{tabular}[t]{l}0.144\end{tabular}}}}%
    \put(0,0){\includegraphics[width=\unitlength,page=18]{mean_indep_example.pdf}}%
    \put(0.85906011,0.36361024){\makebox(0,0)[lt]{\lineheight{1.25}\smash{\begin{tabular}[t]{l}0.192\end{tabular}}}}%
    \put(0,0){\includegraphics[width=\unitlength,page=19]{mean_indep_example.pdf}}%
    \put(0.85906011,0.43028904){\makebox(0,0)[lt]{\lineheight{1.25}\smash{\begin{tabular}[t]{l}0.240\end{tabular}}}}%
    \put(0,0){\includegraphics[width=\unitlength,page=20]{mean_indep_example.pdf}}%
    \put(0.85906011,0.49696783){\makebox(0,0)[lt]{\lineheight{1.25}\smash{\begin{tabular}[t]{l}0.288\end{tabular}}}}%
    \put(0,0){\includegraphics[width=\unitlength,page=21]{mean_indep_example.pdf}}%
    \put(0.85906011,0.56364662){\makebox(0,0)[lt]{\lineheight{1.25}\smash{\begin{tabular}[t]{l}0.336\end{tabular}}}}%
    \put(0,0){\includegraphics[width=\unitlength,page=22]{mean_indep_example.pdf}}%
    \put(0.85906011,0.63032541){\makebox(0,0)[lt]{\lineheight{1.25}\smash{\begin{tabular}[t]{l}0.384\end{tabular}}}}%
    \put(0.97358155,0.38553943){\rotatebox{90}{\makebox(0,0)[t]{\lineheight{1.25}\smash{\begin{tabular}[t]{c}Probability Density\end{tabular}}}}}%
    \put(0,0){\includegraphics[width=\unitlength,page=23]{mean_indep_example.pdf}}%
  \end{picture}%
\endgroup%

%% file: appendix/algo.tex
\section{Algorithm} \label{app:algo}

The complete training procedure for \ADM{} is summarized in \Cref{alg:adversarialgame}.

\begin{algorithm}
\caption{Training Algorithm for Adversarial Dependence Minimization (\ADM{})}
\label{alg:adversarialgame}
\begin{algorithmic}[1]
\Require Encoder $f_\theta$; probe networks $\{\psi_j\}_{j=1}^d$ with parameters $\vpsi$;
         dependency predictors $\{\phi_j\}_{j=1}^d$ with parameters $\vphi$;
         number of critic steps $k$; minibatch size $n$

\For{number of training iterations}

    \vspace{2pt}
    \Comment{\emph{Update the dependence branch $(\vphi,\,\vpsi)$: minimize prediction error}}
    \For{$k$ steps}
        \State Sample a minibatch $\{\vx^{(1)}, \dots, \vx^{(n)}\}$ from the dataset
        \State Compute representations $\vz^{(i)} \gets f_\theta(\vx^{(i)})$ for all $i$ \hfill \Comment{stop gradient w.r.t.\ $\theta$}
        \State Compute batch statistics: $\mu_j = \tfrac{1}{n}\sum_i z_j^{(i)}$,\quad
               $\sigma_j = \sqrt{\tfrac{1}{n-1}\sum_i (z_j^{(i)} - \mu_j)^2}$ \quad for all $j$
        \State Standardize representations: $z_j^{(i)} \gets (z_j^{(i)} - \mu_j)\,/\,\sigma_j$ \quad for all $i,j$
        \State Apply probe networks: $\tilde{z}_j^{(i)} = \psi_j(z_j^{(i)})$ \quad for all $i,j$ 
        \State Compute predictions: $\hat{z}_j^{(i)} = \phi_j\!\left(z_1^{(i)}, \dots, z_{j-1}^{(i)}, z_{j+1}^{(i)}, \dots, z_d^{(i)}\right)$ \quad for all $i,j$
        \State Standardize probe outputs: $\tilde{z}_j^{(i)} \gets \bigl(\tilde{z}_j^{(i)} - \bar{\tilde{z}}_j\bigr)\,/\,\tilde{\sigma}_j$ \quad for all $i,j$
        \State Compute loss $\displaystyle\mathcal{L}_{\mathrm{dep}} = \frac{1}{nd}\sum_{i=1}^n\sum_{j=1}^{d}\Bigl(\tilde{z}_j^{(i)} - \hat{z}_j^{(i)}\Bigr)^{\!2}$
        \State Update $(\vphi,\,\vpsi)$ by gradient descent on $\mathcal{L}_{\mathrm{dep}}$
    \EndFor

    \vspace{2pt}
    \Comment{\emph{Update the encoder $\theta$: maximize prediction error}}
    \State Sample a minibatch $\{\vx^{(1)}, \dots, \vx^{(n)}\}$ from the dataset
    \State Compute representations $\vz^{(i)} \gets f_\theta(\vx^{(i)})$ for all $i$
    \State Compute batch statistics $\mu_j$, $\sigma_j$ as above \quad for all $j$
    \State Standardize representations: $z_j^{(i)} \gets (z_j^{(i)} - \mu_j)\,/\,\sigma_j$ \quad for all $i,j$
    \State Apply probe networks: $\tilde{z}_j^{(i)} = \psi_j(z_j^{(i)})$ \quad for all $i,j$
    \State Compute predictions: $\hat{z}_j^{(i)} = \phi_j\!\left(\vz_{-j}^{(i)}\right)$ \quad for all $i,j$ \hfill\Comment{stop gradient w.r.t.\ $\vphi,\vpsi$}
    \State Standardize probe outputs: $\tilde{z}_j^{(i)} \gets \bigl(\tilde{z}_j^{(i)} - \bar{\tilde{z}}_j\bigr)\,/\,\tilde{\sigma}_j$ \quad for all $i,j$
    \State Compute loss $\displaystyle\mathcal{L}_{\mathrm{adm}} = \frac{1}{nd}\sum_{i=1}^n\sum_{j=1}^{d}1 - \Bigl(\tilde{z}_j^{(i)} - \hat{z}_j^{(i)}\Bigr)^{\!2}$
    \State Update $\theta$ by gradient descent on $\mathcal{L}_{\mathrm{adm}}$
\EndFor
\end{algorithmic}
\end{algorithm}

%% file: appendix/ablation_studies.tex
\section{Ablation Studies}

\subsection{Dependency Predictors' Complexity}

We investigate the impact of the architectural capacity of dependency predictors, specifically their width and depth, on the resulting decorrelation. We follow the same experimental setting as in the convergence analysis on TinyImageNet (Section~\ref{sec:results_convergence}).

\Cref{tab:ablation_width_depth_convergence} reports correlation metrics on the validation set. Increasing the hidden dimension consistently reduces both correlation measures. Notably, even relatively small hidden sizes (e.g., 4) already yield very low correlation levels, suggesting that strong decorrelation can be achieved with minimal predictor capacity.

In contrast, increasing depth does not lead to consistent improvements. One possible explanation is that deeper predictors may be more difficult to optimize under the same training regime, although we leave a more detailed investigation of optimization dynamics to future work. 

\begin{table}[h]
    \small
    \centering
    \caption{Impact of the width and depth of dependency predictors on correlation for TinyImageNet. Both correlation metrics are computed on the validation set. $\dCor$ is estimated on a subset of 5,000 samples using the first 64 dimensions.}
    \label{tab:ablation_width_depth_convergence}
    \begin{tabular}{llrr}
        \toprule
        depth & hidden & $\Cor$ & $\dCor$ \\ \midrule
        2 & 4 & 0.0095 & 0.0354 \\
        2 & 8 & 0.0092 & 0.0294 \\
        2 & 16 & 0.0091 & 0.0276 \\
        2 & 32 & 0.0085 & 0.0188 \\
        2 & 64 & 0.0083 & 0.0148 \\
        2 & 128 & 0.0082 & 0.0146 \\
        \midrule
        3 & 32 & 0.0090 & 0.0257 \\
        \bottomrule
    \end{tabular}
\end{table}

%% file: appendix/additional_results.tex
\section{Nearest Neighbors Visualization} \label{app:visualization}

\begin{figure} %
    \centering
    \includegraphics[width=0.99\textwidth]{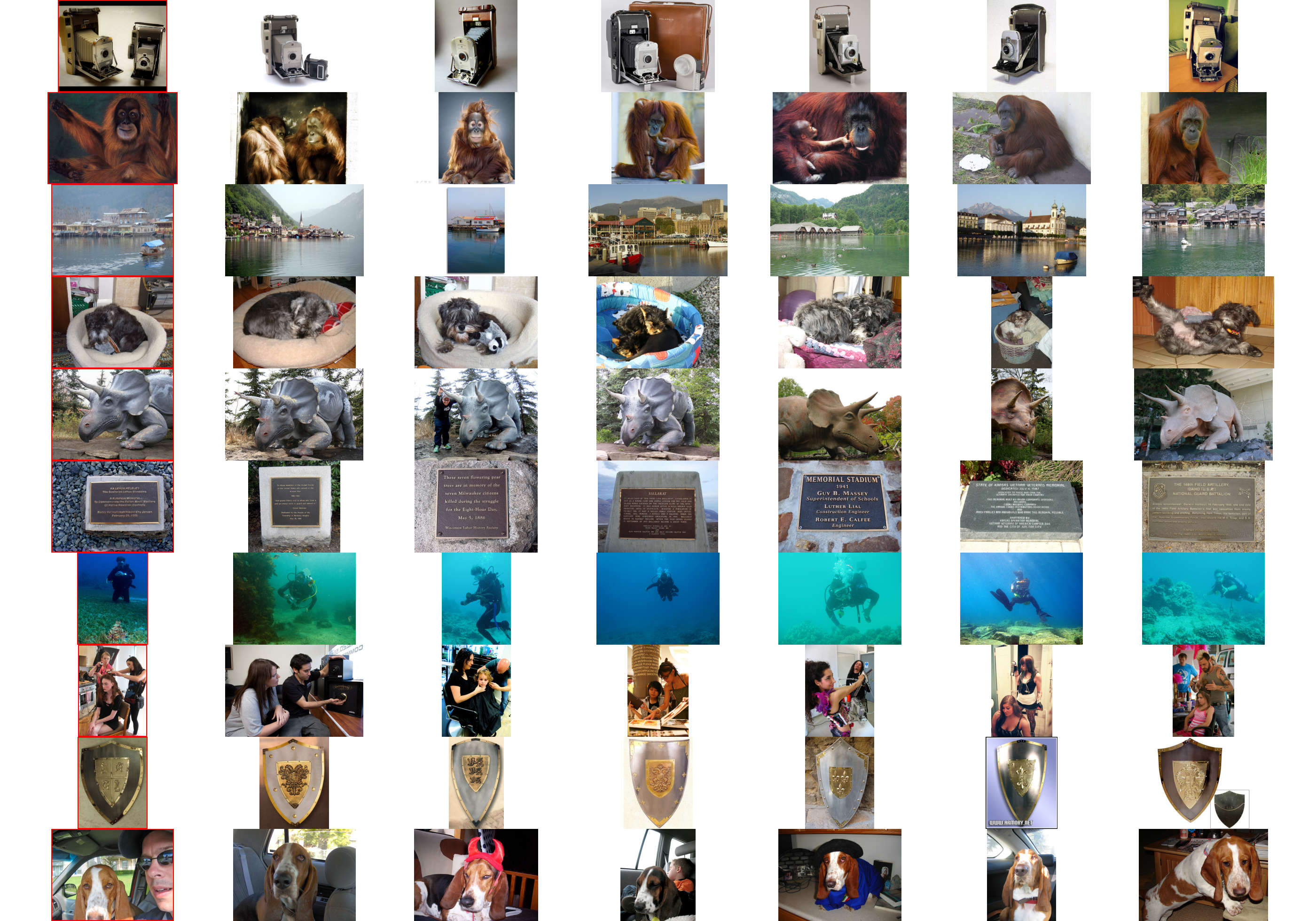}
    \caption{Nearest neighbors visualization for our SSL model trained on the ImageNet dataset. The nearest neighbors visually resemble the query images (highlighted in red).}
    \label{fig:nearest_neighbors_ssl}
\end{figure}

We visualize the nearest neighbors for the self-supervised approach described in \Cref{subsec:ssl_application}. Figure~\ref{fig:nearest_neighbors_ssl} shows the predicted nearest neighbors for 10 randomly sampled validation images from the ImageNet-1k dataset. The left-most image is the query image, and its nearest neighbors are the training samples whose representations have the highest cosine similarity to the query's representation. 
The figure demonstrates that the nearest neighbors visually resemble the query images. 

%% file: appendix/training_details.tex
\section{Detailed Experimental Setups} \label{app:detailed_setup}

We provide here a detailed description of the training settings and hyperparameters to facilitate the reproducibility of our experimental results.

The encoder and dependence branch are trained alternately, following the algorithm presented in Appendix~\ref{app:algo}. Epochs are counted relative to the encoder, so the dependence branch loops through the dataset $k$ times per encoder epoch. 

\paragraph{Dependence branch.} The dependence branch is always trained with the same optimizer and schedulers as the encoder. The default dependency predictor is a two-layer fully connected network with a hidden dimension of 32 and an intermediate GELU~\citep{hendrycks2016gelu} activation function. There is no activation function at the network's output.
Similarly, the default probe network is a two-layer fully connected network with a hidden dimension of 16 and an intermediate GELU activation. 

\paragraph{Variance regularization.} We set the standard deviation hinge loss coefficient to a value of 2 in all experiments, as we found the exact hyperparameter value to have minimal impact. 

\subsection{Experimental Setup: Convergence Analysis} \label{subapp:setup_convergence}

The encoder is a ResNet-18 backbone with no projection head. We used the SGD optimizer with momentum 0.9, learning rate 0.8, batch size 256, and no weight decay. The learning rate follows a cosine decay schedule~\citep{loshchilov2016sgdr} with 10 epochs of linear warmup. The dependence branch is trained with a ratio of $k=16$ steps with a learning rate of 1.6. 
The model is trained on the TinyImageNet dataset for 100 epochs without data augmentation, and images are normalized using ImageNet mean and per-channel standard deviation.

\paragraph{Decorrelation baseline.} The baseline is trained with the same strategy, except the \ADM{} loss is replaced by a simple decorrelation loss that penalizes correlations between feature dimensions, similarly to the approach from BarlowTwins \cite{zbontar2021barlow_ssl}.
More specifically, given a batch of features $\vz \in \mathbb{R}^{b \times d}$, we first standardize each dimension, then compute the empirical covariance matrix $C = \vz^\top \vz / b$ and minimize the mean squared value of its off-diagonal entries:
\[
\mathcal{L}_{\text{decorr}} = \frac{1}{d(d-1)} \sum_{i \neq j} C_{ij}^2.
\]
This objective encourages the learned representation to have an approximately diagonal covariance structure.

\subsection{Experimental Setup: Clevr-4} \label{subapp:setup_clevr4}

\paragraph{Data augmentations.}
We apply two data augmentations during training: random horizontal flipping and random cropping. For random cropping, we keep at least 60\% of the image area, and then resize the images to 224 $\times$ 224 pixels. We do not apply color alterations, as \textit{color} is one of the taxonomies. 

\paragraph{Training.} The adversarial approach is trained with two-layer dependency predictors and probes. The encoder is a ResNet-18 backbone with no projection head. 
We used the SGD optimizer with momentum 0.9, learning rate 0.2, batch size 256, and weight decay $2 \cdot 10^{-5}$. 
The learning rate follows a cosine decay schedule~\citep{loshchilov2016sgdr} with 10 epochs of linear warmup. The dependence networks are trained with a ratio of $k=1$ steps with a learning rate of 0.8. 
The baseline classifier follows the same strategy, except it only uses a cross-entropy loss.

\subsection{Experimental Setup: ImageNet SSL} \label{subapp:setup_imagenet_ssl}

We trained two-layer dependency predictors with no probes. The encoder is a ResNet-50 backbone with a three-layer fully connected projection head with a hidden dimension of 4096, an output dimension of 512, ReLU activations, and intermediate BatchNorm layers. 
We used the LARS optimizer~\citep{you2017LARS} with a momentum of 0.9, a base learning rate of 1.5 with linear scaling rule~\citep{goyal2017lr_scaling_rule}, a batch size of 1024, and a weight decay of $10^{-4}$. 
The learning rate follows a cosine decay schedule~\citep{loshchilov2016sgdr} with 10 epochs of linear warmup. The dependence branch is trained with a ratio of $k=4$ steps with a base learning rate of 6. 
The model is trained on the ImageNet dataset for 300 epochs and follows the same data augmentations as in~\cite{grill2020BYOL_ssl}. 

\paragraph{Linear evaluation.} We followed standard procedure and trained a linear classifier on top of the frozen representations from the backbone. We used the SGD optimizer with a learning rate of 1.5, a weight decay of $10^{-6}$, a batch size of 256, and trained for 100 epochs. The learning rate follows a cosine decay schedule~\citep{loshchilov2016sgdr}. 
We applied two data augmentations during training: random horizontal flipping with $p=0.5$ and random cropping by keeping at least 8\% of the image area, followed by resizing to 224 $\times$ 224 pixels. 
During the evaluation, the images were resized so that the shorter side was 256 pixels, then center-cropped to 224 $\times$ 224 pixels.